\documentclass[10pt,journal,compsoc]{IEEEtran}
\usepackage{xcolor}
\usepackage{amsfonts, amsmath, amssymb, amsthm}

\theoremstyle{definition}
\newtheorem{remark}{Remark}

\usepackage{graphicx}
\usepackage{subfigure}
\usepackage{subcaption}
\usepackage{float}
\usepackage{tabularx}
\usepackage{booktabs}
\usepackage{multicol}
\usepackage{multirow}
\PassOptionsToPackage{hyphens}{url}\usepackage[colorlinks]{hyperref}
\hypersetup{colorlinks,breaklinks,linkcolor=red,urlcolor=blue,anchorcolor=blue,citecolor=blue}
\usepackage{algorithm}
\usepackage{algorithmic}

\DeclareMathOperator*{\argmin}{arg\,min}

\ifCLASSOPTIONcompsoc
  \usepackage[nocompress]{cite}
\else
  \usepackage{cite}
\fi

\ifCLASSINFOpdf
\else
\fi

\begin{document}
%

\title{Interpretable Time Series Autoregression for Periodicity Quantification
}
%
%
%
%

\author{
Xinyu~Chen, Vassilis~Digalakis~Jr, Lijun~Ding, Dingyi~Zhuang, and~Jinhua Zhao

\IEEEcompsocitemizethanks{\IEEEcompsocthanksitem Xinyu Chen, Dingyi Zhuang, and Jinhua Zhao are with the Department of Urban Studies and Planning, Massachusetts Institute of Technology, Cambridge, MA 02139, USA (e-mail: chenxy346@gmail.com; dingyi@mit.edu; jinhua@mit.edu).
\IEEEcompsocthanksitem Vassilis Digalakis Jr is with the Questrom School of Business, Boston University, Boston, MA 02115, USA (e-mail: vvdigalakis@gmail.com).
\IEEEcompsocthanksitem Lijun Ding is with the Department of Mathematics at the University of California, San Diego, La Jolla, CA 92093, USA (e-mail: l2ding@ucsd.edu).

}
\thanks{(Corresponding author: Jinhua Zhao)}}

\IEEEtitleabstractindextext{%
\begin{abstract}

Time series autoregression (AR) is a classical tool for modeling auto-correlations and periodic structures in real-world systems. We revisit this model from an interpretable machine learning perspective by introducing sparse autoregression (SAR), where $\ell_0$-norm constraints are used to isolate dominant periodicities. We formulate exact mixed-integer optimization (MIO) approaches for both stationary and non-stationary settings and introduce two scalable extensions: a decision variable pruning (DVP) strategy for temporally-varying SAR (TV-SAR), and a two-stage optimization scheme for spatially- and temporally-varying SAR (STV-SAR). These models enable scalable inference on real-world spatiotemporal datasets. We validate our framework on large-scale mobility and climate time series. On NYC ridesharing data, TV-SAR reveals interpretable daily and weekly cycles as well as long-term shifts due to COVID-19. On climate datasets, STV-SAR uncovers the evolving spatial structure of temperature and precipitation seasonality across four decades in North America and detects global sea surface temperature dynamics, including El Niño. Together, our results demonstrate the interpretability, flexibility, and scalability of sparse autoregression for periodicity quantification in complex time series.

\end{abstract}

}

\maketitle

\IEEEdisplaynontitleabstractindextext

%
\IEEEpeerreviewmaketitle

\section{Introduction}
\IEEEPARstart{M}{any} real-world systems exhibit complex temporal patterns, including periodicity, seasonality, and anomalies. Detecting and understanding these patterns is essential for anticipating system behavior, identifying disruptions, and supporting operational decision-making. In dynamic environments such as urban transportation and climate systems, periodicities can shift due to external factors—policy interventions, demand changes, environmental variability, global events, or extreme climate phenomena—making interpretable, adaptive, and data-driven methods indispensable. A central challenge is to automatically identify dominant periodic components from temporally- and spatially-varying systems, track their evolution over a long-term time period, and distinguish true structural changes from random variability.

\begin{figure}[ht!]
\centering
\includegraphics[width=0.45\textwidth]{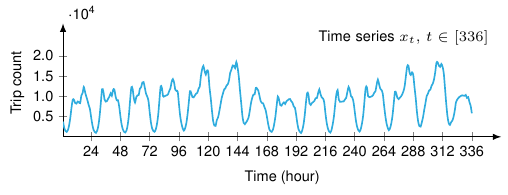}
\caption{Hourly time series of aggregated ridesharing trip counts in Chicago during the first two weeks (336 hours) starting April 1, 2024. The data exhibits strong periodicity with a weekly cycle $\Delta t = 7 \times 24 = 168$.}
\label{chicago_rideshare_ts}
\end{figure}

\textbf{Urban transportation systems} display strong periodicity driven by commuting patterns, business cycles, and travel demand. Figure~\ref{chicago_rideshare_ts} shows the regularity of ridesharing activity in Chicago, highlighting its weekly mobility patterns. However, such patterns are not fixed—they evolve with infrastructure changes, economic conditions, and disruptive events. For example, the COVID-19 pandemic in 2020 led to a collapse in established mobility periodicities due to lockdowns and shifts to remote work. These disruptions raise key questions: \emph{How do periodic structures evolve over time? Can we systematically quantify such changes in a temporally-varying system?} Addressing these questions is critical for planning, forecasting, and adaptive resource allocation.

\textbf{Climate systems} also exhibit periodic and seasonal patterns that shape temperature, ocean circulation, and atmospheric dynamics. Yet these patterns evolve over time due to long-term variability and climate change. Traditional time series decomposition methods \cite{pezzulli2005variability, feng2013changes, tonkin2017seasonality} assume fixed seasonal structure and often ignore gradual or region-specific shifts. Accurately monitoring such changes is essential for understanding climate dynamics, anticipating extreme events, and supporting policy-making. This calls for \emph{interpretable, data-driven models that can robustly uncover the dominant seasonal components and track their evolution across space and time}.

To address these needs, we develop a unified interpretable machine learning framework for identifying and quantifying temporally-varying periodicity and structural shifts in real-world time series. Our models build on classical autoregression (AR) \cite{box2015time, hamilton2020time} but incorporate sparse structure via $\ell_0$-norm induced sparsity constraints to promote interpretability. Inspired by recent advances in sparse regression \cite{bertsimas2024slowly} and time series convolution \cite{chen2025correlating}, we formulate interpretable AR models capable of isolating dominant periodic patterns over time and space. To the best of our knowledge, this represents the first application of exact sparse AR to the task of periodicity quantification in real-world complex time series. Overall, this work makes the following contributions:
\begin{itemize}
    \item \textbf{Sparse Autoregression (SAR):} We introduce an interpretable framework for identifying dominant auto-correlations from time series by reformulating AR with $\ell_0$-norm induced sparsity constraints. The problem is solved exactly via mixed-integer optimization (MIO) techniques, providing more accurate and reliable periodicity quantification than conventional greedy methods.

    \item \textbf{Temporally-Varying SAR (TV-SAR):} We extend SAR to characterize non-stationary time series, enforcing consistent support sets across time segments to enhance interpretability. To improve scalability, we introduce a decision variable pruning (DVP) strategy that narrows the MIO search space using fast greedy approximations such as subspace pursuit.

    \item \textbf{Spatially- and Temporally-Varying SAR (STV-SAR):} We propose a scalable model for multidimensional time series that vary over both space and time. A two-stage optimization procedure—global support set selection via MIO, followed by local coefficient estimation via quadratic optimization—makes the method tractable for millions of decision variables.

    \item \textbf{Extensive Real-World Validation:} We demonstrate the effectiveness of our models on large-scale transportation and climate datasets. TV-SAR reveals dynamic changes in daily and weekly mobility patterns related to periodicity in New York City (NYC) during the period of the COVID-19 pandemic. STV-SAR captures evolving spatial patterns of seasonality in North American climate variables and identifies global sea surface temperature dynamics related to El Niño.
\end{itemize}

The remainder of this paper is organized as follows. Section~\ref{literature} reviews related work. Section~\ref{preliminaries} introduces notation and background on AR models. Section~\ref{sparse_autoregression} presents the core SAR model. Sections~\ref{time_varying_model} and~\ref{spatially_varying_model} develop the TV-SAR and STV-SAR extensions, respectively. Section~\ref{experiments} reports empirical results. Finally, we conclude this study in Section~\ref{conclusion}.

\section{Literature Review}\label{literature}

\subsection{Classical and Temporally-Varying Autoregression}

Time series modeling is a foundational tool in transportation, climate science, econometrics, and other fields \cite{chen2024discovering, kutz2016dynamic, cochrane1997time}. 
Classical models such as autoregression (AR), moving average (MA), and their combinations (e.g., ARIMA) are widely used for capturing temporal dependencies and seasonal structures in univariate time series data \cite{box2015time, hamilton2020time,brockwell1991time}. 
Among these, AR models remain popular due to their simplicity and interpretability. More recently, applications have demanded models that adapt to non-stationary and temporally-varying systems.
Temporally-varying AR extends classical AR by allowing coefficients to change over time \cite{haslbeck2021tutorial}, often incorporating structural or smoothness constraints to ensure interpretability and stability \cite{harris2021time}. AR has also been generalized to multivariate and spatial-temporal settings, such as Vector Autoregressive (VAR) models \cite{bringmann2017changing,primiceri2005time}. 
These ideas have been successfully applied in finance, neuroscience, and dynamic mobility systems.

\subsection{Sparse Autoregression and Interpretability}
Traditional AR models include all lagged terms up to a fixed order, which can lead to overfitting and obscure the most meaningful temporal dependencies—particularly in high-order settings \cite{nardi2011autoregressive}. 
Sparse AR addresses this limitation by selecting a subset of informative lags, thereby improving both parsimony and interpretability. LASSO-based methods \cite{tibshirani1996regression} have been widely used to induce sparsity in AR models, offering scalability and robustness in high-dimensional contexts \cite{davis2016sparse}. Recent work has further explored structured sparsity in time series models. For instance, \cite{brunton2016discovering, bertsimas2023learning, liu2024okridge} develop interpretable sparse formulations for nonlinear and dynamical systems, while \cite{bertsimas2024slowly} introduces structured sparsity constraints over graphs to model slowly evolving regression coefficients. 
A related line of research by \cite{carrizosa2017sparsity} proposes a sparsity-controlled VAR framework, allowing users to tune multiple dimensions of sparsity for enhanced interpretability of causal discovery. 
While this approach offers flexibility and shows improved predictive accuracy over LASSO-based alternatives, it lacks scalability and does not provide exact solutions with guaranteed optimality.

These advances move beyond classical LASSO penalties by incorporating domain-specific sparsity structures. However, existing methods typically do not enforce structured sparsity across time and space simultaneously, nor do they leverage exact combinatorial optimization. Our work bridges this gap by introducing a framework for structured sparse AR with exact support set control over spatiotemporal dimensions.

\subsection{Exact Sparse Regression via Mixed-Integer Optimization}

Sparse regression via $\ell_0$-norm regularization—also known as best subset selection—has long been recognized for its statistical optimality and interpretability, but was historically limited by its combinatorial complexity. Early heuristic methods include Orthogonal Matching Pursuit and CoSaMP \cite{pati1993orthogonal, needell2009cosamp, dai2009subspace}, while convex relaxations such as the LASSO \cite{tibshirani1996regression} and non-convex penalties such as SCAD and MCP offered tractable alternatives \cite{wang2020simple}. The solution quality of these methods often negatively impact the interpretability of sparse structures. Recent breakthroughs in optimization have made it practically feasible to solve $\ell_0$-norm regularized problems exactly using MIO. Following the seminal work of \cite{bertsimas2016best}, several studies have scaled MIO-based sparse regression to large datasets \cite{bertsimas2020sparse, hazimeh2022sparse, atamturk2025rank}. These methods retain the interpretability of exact sparse models while achieving near-LASSO speed. 

Our work builds directly on this line, leveraging the MIO machinery developed in \cite{bertsimas2024slowly} to solve structured sparse regression with controlled support set consistency. However, unlike prior work which focused on static or graph-based regression, we apply these methods to time series with dynamic and spatial variation, developing the first exact sparse AR framework for periodicity quantification in large-scale real-world systems.

\section{Preliminaries}\label{preliminaries}

\subsection{Notation}

In Table~\ref{notation}, we summarize the basic symbols and notation used throughout the paper. Notably, $\mathbb{R}$ denotes the set of real numbers, and $\mathbb{Z}^{+}$ denotes the set of positive integers.

\begin{table}[h!]
\centering
\caption{Summary of basic notation.}
\label{notation}
\begin{tabular}{l|l} 
\toprule
Notation & Description \\ 
\midrule
$x \in \mathbb{R}$ & Scalar \\
$\boldsymbol{x} \in \mathbb{R}^{n}$ & Vector of length $n$ \\
$\boldsymbol{X} \in \mathbb{R}^{m \times n}$ & Matrix of size $m \times n$ \\
$[i]$ & Integer set $\{1, 2, \ldots, i\}$, $i \in \mathbb{Z}^{+}$ \\
$[i, j]$ & Integer set $\{i, i+1, \ldots, j\}$, $i < j$ \\
$\|\cdot\|_0$ & $\ell_0$-norm (number of nonzero entries) \\
$\|\cdot\|_1$ & $\ell_1$-norm (sum of absolute values) \\
$\|\cdot\|_2$ & $\ell_2$-norm (Euclidean norm) \\
$\operatorname{tr}(\cdot)$ & Trace of a square matrix \\
$\operatorname{supp}(\cdot)$ & Support set (indices of nonzero entries) \\
$\mathcal{N}(\cdot)$ & Gaussian distribution \\
$\mathbb{E}[\cdot]$ & Expectation \\
$\cup$ & Union of sets \\
$\cap$ & Intersection of sets \\
\bottomrule
\end{tabular}
\end{table}

\subsection{Time Series Autoregression}

AR is a widely used technique for modeling temporal dependencies in univariate time series \cite{box2015time, hamilton2020time}. It expresses each observation as a linear combination of its past values, plus noise. For a univariate time series $\boldsymbol{x} = (x_1, x_2, \ldots, x_T)^\top \in \mathbb{R}^T$, the order-$d$ AR model is written as:
\begin{equation}\label{time_series_autoregression}
x_t = \sum_{k=1}^{d} w_k x_{t-k} + \epsilon_t, \quad \forall t \in [d+1, T],
\end{equation}
where $d \in \mathbb{Z}^{+}$ is the AR order, and $\boldsymbol{w} = (w_1, w_2, \ldots, w_d)^\top \in \mathbb{R}^d$ is the coefficient vector. As the residual, $\epsilon_t$ denotes noise, typically modeled by a Gaussian assumption: $\epsilon_t \sim \mathcal{N}(0, \sigma^2)$ for the variance $\sigma^2>0$. The coefficient $w_k$ captures the linear dependence between $x_t$ and its $k$-lagged value $x_{t-k}$.

To estimate the coefficients, we minimize the sum of squared residuals such that
\begin{equation}\label{time_series_autoregression_opt}
\hat{\boldsymbol{w}} = \argmin_{\boldsymbol{w}} \sum_{t=d+1}^{T} \left( x_t - \sum_{k=1}^{d} w_k x_{t-k} \right)^2.
\end{equation}
By defining the $(T-d)$-by-$d$ design matrix and length-$(T-d)$ target vector as:
\begin{equation}\label{data_pair}
\boldsymbol{A} = 
\begin{bmatrix}
x_d & x_{d-1} & \cdots & x_1 \\
x_{d+1} & x_d & \cdots & x_2 \\
\vdots & \vdots & \ddots & \vdots \\
x_{T-1} & x_{T-2} & \cdots & x_{T-d}
\end{bmatrix}, \quad
\tilde{\boldsymbol{x}} = 
\begin{bmatrix}
x_{d+1} \\ x_{d+2} \\ \vdots \\ x_T
\end{bmatrix},
\end{equation}
respectively. Then, Problem \eqref{time_series_autoregression_opt} becomes
\[
\hat{\boldsymbol{w}} = \argmin_{\boldsymbol{w}} \|\tilde{\boldsymbol{x}} - \boldsymbol{A}\boldsymbol{w}\|_2^2.
\]
This leads to the standard least squares solution:
\(
\hat{\boldsymbol{w}} = (\boldsymbol{A}^\top \boldsymbol{A})^{-1} \boldsymbol{A}^\top \tilde{\boldsymbol{x}},
\)
assuming $\boldsymbol{A}^\top \boldsymbol{A}$ is invertible. In more general settings, this solution can also be expressed using the Moore-Penrose pseudoinverse as $\hat{\boldsymbol{w}} = \boldsymbol{A}^{\dagger} \tilde{\boldsymbol{x}}$.

A time series $\boldsymbol{x}$ is said to exhibit periodicity with period $\Delta t \in \mathbb{Z}^+$ if $x_t \approx x_{t + \Delta t}$ for many $t$, typically reflected by high auto-correlation at lag $\Delta t$, i.e., $\operatorname{Cov}(x_t, x_{t + \Delta t}) \gg 0$. Seasonality refers to periodic patterns tied to calendar cycles (e.g., daily, weekly, and yearly) and can be modeled as a component $s_t$ satisfying $s_t = s_{t + \Delta t}$. In the AR framework, strong periodicity at lag $\Delta t$ manifests as a large positive coefficient $w_{\Delta t}$, allowing us to infer dominant cycles directly from the estimated sparse coefficient vector $\boldsymbol{w}$.

\begin{figure}[ht!]
\centering
\includegraphics[width=0.45\textwidth]{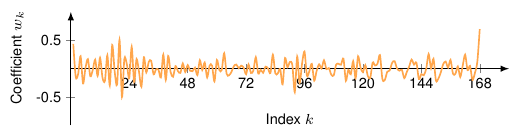}
\caption{Estimated AR coefficients $\boldsymbol{w} \in \mathbb{R}^{d}$ for the Chicago ridesharing time series in Figure~\ref{chicago_rideshare_ts}, with order $d = 168$.}
\label{chicago_rideshare_coefficients}
\end{figure}

Figure~\ref{chicago_rideshare_coefficients} illustrates the coefficient vector $\boldsymbol{w}$ estimated for the Chicago ridesharing time series using standard least squares. The resulting vector is dense, with both positive and negative coefficients. A large positive $w_k$ at lag $k = 168$ reflects the strong weekly periodicity of the time series. However, this dense representation makes it difficult to identify which lags are most important and to quantify periodic structure precisely. This motivates the need for sparse and interpretable AR models.

\section{Sparse Autoregression}\label{sparse_autoregression}

This section introduces Sparse Autoregression (SAR) with $\ell_0$-norm induced sparsity constraints for identifying dominant auto-correlations in time series. We first describe the modeling framework; then, we present an MIO formulation for solving the resulting optimization problem exactly and compare solution quality across algorithms.

\subsection{Model Description} 
The $\ell_0$-norm of a vector $\boldsymbol{w} \in \mathbb{R}^d$ is defined as $\|\boldsymbol{w}\|_0 = |\operatorname{supp}(\boldsymbol{w})| \leq d$, i.e., the number of nonzero entries. While the least squares estimator in Eq.~\eqref{time_series_autoregression_opt} produces dense solutions, it does not highlight dominant auto-correlations, making it difficult to quantify periodicity or seasonality.

To address this, we impose sparsity and non-negativity constraints on the coefficient vector $\boldsymbol{w}$, yielding the following optimization problem:
\begin{equation}\label{optimization_sparse_autoregression}
\begin{aligned}
\min_{\boldsymbol{w}} \quad & \|\tilde{\boldsymbol{x}} - \boldsymbol{A}\boldsymbol{w}\|_2^2 \\
\text{s.t.} \quad & 0 \leq \boldsymbol{w} \leq \boldsymbol{\mathcal{M}}, \quad \|\boldsymbol{w}\|_0 \leq \tau,
\end{aligned}
\end{equation}
where $\boldsymbol{\mathcal{M}} \in \mathbb{R}^d$ is a vector with all entries equal to a sufficiently large constant $\mathcal{M} \in \mathbb{R}$, and $\tau \in \mathbb{Z}^+$ controls the maximum number of nonzero coefficients. The non-negativity constraint encourages interpretability by focusing on positive auto-correlations, as is typical for periodicity in time series. 

\subsection{MIO Reformulation}
Due to the combinatorial nature of the $\ell_0$-norm induced constraint, Problem~\eqref{optimization_sparse_autoregression} cannot be directly solved by MIO solvers. However, we can equivalently express the problem as a mixed-integer quadratic optimization problem by introducing binary variables that encode the support of $\boldsymbol{w}$.
Let $\boldsymbol{z} \in \{0,1\}^d$ be a vector of binary variables, where $z_k = 1$ indicates that $w_k$ is allowed to be nonzero. We rewrite Problem~\eqref{optimization_sparse_autoregression} as:
\begin{equation} \label{eq:miqp_sparse_autoreg}
\begin{aligned}
\min_{\boldsymbol{w}, \boldsymbol{z}} \quad & \|\tilde{\boldsymbol{x}} - \boldsymbol{A} \boldsymbol{w}\|_2^2 \\
\text{s.t.} \quad & 0 \leq w_k \leq \mathcal{M}\cdot z_k, \quad \forall k \in [d], \\
& \sum_{k=1}^d z_k \leq \tau, \\
& z_k \in \{0,1\}, \quad \forall k \in [d].
\end{aligned}
\end{equation}
The binary support constraint $\displaystyle\sum_{k=1}^d z_k \leq \tau$ ensures that at most $\tau$ coefficients in $\boldsymbol{w}$ are nonzero. 

Problem~\eqref{eq:miqp_sparse_autoreg} is nonconvex due to the $\ell_0$-norm induced constraint and is generally NP-hard to solve exactly. In principle, enumerating all support sets of cardinality $\tau$ would incur a combinatorial cost of $\mathcal{O}(d^\tau)$, making exact search infeasible for large $d$ or $\tau$. Modern MIO solvers bypass this challenge using branch-and-bound techniques and cutting-plane methods (e.g., \cite{bertsimas2024slowly}), which allow them to find globally optimal solutions efficiently in practice. Compared to greedy methods such as subspace pursuit, MIO yields higher-quality solutions with provable optimality guarantees. The use of binary variables enables precise support set control, which is essential for model interpretability in settings where dominant lag selection matters. 

\subsection{Empirical Comparison of Solution Quality}
The estimation method has a significant impact on the interpretability and fidelity of SAR models. To highlight this, we compare two approaches for solving Problem~\eqref{optimization_sparse_autoregression}: (i) a greedy non-negative subspace pursuit (NNSP) algorithm \cite{chen2025correlating}, and (ii) the exact MIO formulation from Eq.~\eqref{eq:miqp_sparse_autoreg}. We use the time series from Figure~\ref{chicago_rideshare_ts}, with AR order $d = 168$ (one week of hourly lags) and sparsity level $\tau = 2$. The solution returned by NNSP is:
\[
\boldsymbol{w} = (0,\dots,0,\underbrace{0.02}_{k=53},0,\dots,0,\underbrace{0.96}_{k=168})^\top,
\]
with objective value $f(\boldsymbol{w}) = 8.32 \times 10^7$. In contrast, the MIO solver yields:
\[
\boldsymbol{w} = (\underbrace{0.22}_{k=1},0,\dots,0,\underbrace{0.77}_{k=168})^\top,
\]
with objective value $f(\boldsymbol{w}) = 6.25 \times 10^7$. The MIO solution is both quantitatively superior (lower error) and qualitatively more interpretable: it identifies lag $k=1$ (local auto-correlation) and lag $k=168$ (weekly periodicity), consistent with domain knowledge. The NNSP solution, by contrast, includes a spurious lag at $k=53$ with negligible weight. As can be seen, the choice of solution algorithms affect not only the objective function of optimization problem but the interpretability of sparse coefficient vector as well.

When increasing the sparsity level to $\tau = 3$, both solvers return the coefficient vector such that
\[
\boldsymbol{w} = (\underbrace{0.33}_{k=1},0,\dots,0,\underbrace{0.20}_{k=167},\underbrace{0.46}_{k=168})^\top,
\]
highlighting strong lags near daily and weekly cycles.

\begin{figure}[ht!]
\centering
\subfigure[Dominant support sets]{
    \centering
    \includegraphics[width = 0.24\textwidth]{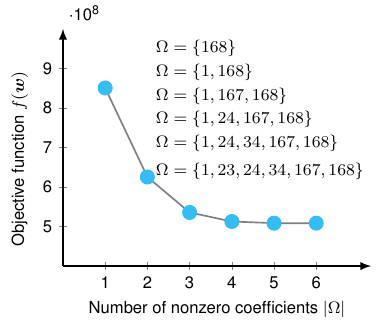}
}\hspace{-1em}
\subfigure[Dominant coefficients ($|\Omega|=5$)]{
    \centering
    \includegraphics[width = 0.24\textwidth]{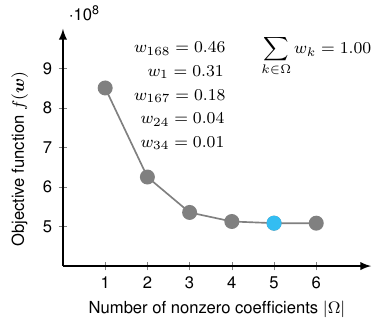}
}
\caption{Illustration of the dominant coefficients of SAR on the Chicago ridesharing trip time series (see Figure~\ref{chicago_rideshare_ts}) at different sparsity levels. The support set and the number of nonzero coefficients are denoted by $\Omega$ and $|\Omega|$, respectively.}
\label{dominant_supp_plus_coef}
\end{figure}

Figure~\ref{dominant_supp_plus_coef} summarizes how the selected support sets and coefficient magnitudes evolve as a function of $\tau$. These results confirm that high-quality solutions are critical for interpretability and that MIO offers robust, principled support recovery even when greedy methods fail.

\section{Temporally-Varying Sparse Autoregression}\label{time_varying_model}

In this section, we now extend the SAR formulation to capture non-stationary dynamics by allowing the coefficients to vary across prescribed time segments. This yields the Temporally-Varying Sparse Autoregression (TV-SAR) model. To solve the associated MIO problem more efficiently, we also introduce a decomposition-based DVP strategy.

\subsection{Model Description}

Temporally-varying AR models allow the AR coefficients to evolve across time, capturing structural shifts in the data-generating process. We partition the time series into $\Gamma \in \mathbb{Z}^+$ time segments, where each segment $\gamma \in [\Gamma]$ contains $T_\gamma \in \mathbb{Z}^+$ time steps. Let $\boldsymbol{x}_{\gamma} = (x_{\gamma,1}, x_{\gamma,2}, \dots, x_{\gamma,T_\gamma})^\top \in \mathbb{R}^{T_\gamma}$ denote the segment corresponding to time segment $\gamma$. We model each segment with an independent $d$th-order AR process, with potentially different coefficient vectors $\boldsymbol{w}_\gamma \in \mathbb{R}^d$:
\[
x_{\gamma,t} = \sum_{k=1}^d w_{\gamma,k} x_{\gamma,t-k} + \epsilon_{\gamma,t}, \quad \forall t \in [d+1, T_\gamma],\, \gamma \in [\Gamma],
\]
where $\epsilon_{\gamma,t} \sim \mathcal{N}(0, \sigma^2)$. To ensure interpretability and temporal consistency, we impose the following structural constraints:
(i) each $\boldsymbol{w}_\gamma$ is sparse and non-negative, and
(ii) all $\boldsymbol{w}_\gamma$ share the same support set. Formally, we let $\Phi=\{\boldsymbol{w}_{\gamma}\}_{\gamma=1}^{\Gamma}$ represent the set of coefficient vectors, referring to the decision variables in the optimization problem of TV-SAR such that
\begin{equation} \label{optimization_time_varying_autoregression}
\begin{aligned}
\min_{\Phi} \quad & \sum_{\gamma=1}^\Gamma \| \tilde{\boldsymbol{x}}_\gamma - \boldsymbol{A}_\gamma \boldsymbol{w}_\gamma \|_2^2 \\
\text{s.t.} \quad &
0 \leq \boldsymbol{w}_\gamma \leq \boldsymbol{\mathcal{M}}, \quad \|\boldsymbol{w}_\gamma\|_0 \leq \tau, \quad \forall \gamma \in [\Gamma], \\
& \operatorname{supp}(\boldsymbol{w}_\gamma) = \operatorname{supp}(\boldsymbol{w}_{\gamma+1}), \quad \forall \gamma \in [\Gamma - 1],
\end{aligned}
\end{equation}
where $\boldsymbol{A}_\gamma \in \mathbb{R}^{(T_\gamma - d) \times d}$ is the design matrix, $\tilde{\boldsymbol{x}}_\gamma \in \mathbb{R}^{T_\gamma - d}$ is the response vector for time segment $\gamma$, $\boldsymbol{\mathcal{M}} \in \mathbb{R}^d$ consists of all sufficiently large constant $\mathcal{M}>0$, and $\tau \in \mathbb{Z}^+$ controls global sparsity.

\begin{remark}
The constraint of the same support set enforces temporal smoothness by requiring identical sparsity patterns across time segments. This is a special case of the sparsely-varying support set constraint proposed in~\cite{bertsimas2024slowly}, where the symmetric difference between support sets is bounded:
\[
|\operatorname{supp}(\boldsymbol{w}_\gamma) \cup \operatorname{supp}(\boldsymbol{w}_{\gamma+1})| - 
|\operatorname{supp}(\boldsymbol{w}_\gamma) \cap \operatorname{supp}(\boldsymbol{w}_{\gamma+1})| \leq \tilde{\tau}.
\]
Setting $\tilde{\tau} = 0$ recovers the shared support set case.
\end{remark}

To encode Eq.~\eqref{optimization_time_varying_autoregression} as an MIO problem, we again introduce binary variables $\boldsymbol{z} \in \{0,1\}^d$ to represent the global support set. The resulting MIO formulation is given by
\begin{equation} \label{time_varying_autoregression_w_mip}
\begin{aligned}
\min_{\Phi,\, \boldsymbol{z}} \quad & \sum_{\gamma=1}^\Gamma \| \tilde{\boldsymbol{x}}_\gamma - \boldsymbol{A}_\gamma \boldsymbol{w}_\gamma \|_2^2 \\
\text{s.t.} \quad &
0 \leq \boldsymbol{w}_\gamma \leq \mathcal{M} \cdot \boldsymbol{z}, \quad \forall \gamma \in [\Gamma], \\
& \sum_{k=1}^d z_k \leq \tau, \\
& z_k \in \{0,1\}, \quad \forall k \in [d].
\end{aligned}
\end{equation}
Here, the binary variable $z_k = 1$ if lag $k$ is selected for any $\gamma$, enforcing a common support set across all time segments. TV-SAR thus extends SAR to non-stationary settings, allowing the autoregressive weights to vary across time segments while preserving interpretability through global sparsity and support set consistency.

\subsection{Acceleration with Decision Variable Pruning}
\begin{figure}[ht!]
\centering
\includegraphics[width=0.48\textwidth]{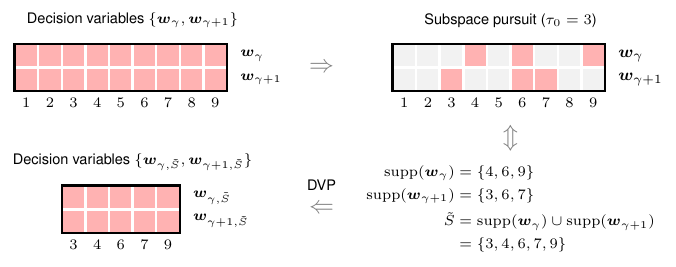}
\caption{Illustration of the DVP strategy. Subspace pursuit is used to select an index set $\tilde{S}$ of candidate lags. The final MIO is then solved on this reduced support set, where the number of coefficients is reduced from $2d$ to $2|\tilde{S}|$ (e.g., from 18 coefficients to 10 coefficients in the illustration).}
\label{opt_strategy}
\end{figure}

The computational cost of solving TV-SAR with MIO increases with the AR order $d$, due to the total number of decision variables $(\Gamma+1)d$, which includes $\Gamma d$ real-valued variables for the AR coefficients $\boldsymbol{w}_\gamma$ and $d$ binary variables in $\boldsymbol{z}$. When the sparsity level $\tau$ is much smaller than $d$, most of these variables are expected to be zero. We exploit this by introducing a \textit{Decision Variable Pruning (DVP)} strategy, which leverages subspace pursuit to pre-select a reduced candidate set of variables before solving the MIO. The strategy proceeds in three steps (illustrated in Figure~\ref{opt_strategy}):
\begin{itemize}
\item \textbf{Step 1: Relax and Decompose TV-SAR.}  
We relax the TV-SAR formulation in Eq.~\eqref{optimization_time_varying_autoregression}, keeping only non-negativity and a looser sparsity constraint $\tau_0 > \tau$. This yields $\Gamma$ decomposable subproblems:
\begin{equation} \label{nnsp_subproblem}
\begin{aligned}
\min_{\boldsymbol{w}_\gamma} \quad & \|\tilde{\boldsymbol{x}}_\gamma - \boldsymbol{A}_\gamma \boldsymbol{w}_\gamma\|_2^2 \\
\text{s.t.} \quad & \boldsymbol{w}_\gamma \geq 0,\quad \|\boldsymbol{w}_\gamma\|_0 \leq \tau_0.
\end{aligned}
\end{equation}

\item \textbf{Step 2: Implement Subspace Pursuit.}  
We first solve each subproblem using NNSP. Let $S_\gamma$ denote the support set of the resulting solution for time segment $\gamma\in[\Gamma]$, see Algorithm~\ref{variable_pruning}. We then form the global candidate set
\[
\tilde{S} = \bigcup_{\gamma=1}^\Gamma S_\gamma.
\]

\item \textbf{Step 3: Solve the Reduced MIO.}  
We solve the original MIO problem in Eq.~\eqref{time_varying_autoregression_w_mip}, restricting all decision variables to the reduced index set $\tilde{S}$. This results in a significantly smaller problem with $(\Gamma+1) \cdot |\tilde{S}|$ variables.
\end{itemize}

\begin{algorithm}[h]
\caption{Decision Variable Pruning via Non-Negative Subspace Pursuit}
\label{variable_pruning}
\begin{algorithmic}[1]
   \STATE \textbf{Input:} Time series $\boldsymbol{x}_\gamma \in \mathbb{R}^{T},\ \gamma \in [\Gamma]$; AR order $d$; relaxed sparsity $\tau_0 > \tau$.
   \FOR{$\gamma \in [\Gamma]$}
       \STATE Construct $\tilde{\boldsymbol{x}}_\gamma \in \mathbb{R}^{T-d}$ and design matrix $\boldsymbol{A}_\gamma \in \mathbb{R}^{(T-d)\times d}$.
       \STATE Initialize $\boldsymbol{w}_\gamma := \boldsymbol{0}$, $S_\gamma := \emptyset$, and residual $\boldsymbol{r} := \tilde{\boldsymbol{x}}_\gamma$.
       \WHILE{not converged}
           \STATE Identify $\ell$: index set of $\tau$ largest entries in $|\boldsymbol{A}_\gamma^\top \boldsymbol{r}|$.
           \STATE $S_\gamma \gets S_\gamma \cup \ell$
           \STATE Solve $\displaystyle\boldsymbol{w}_{\gamma,S_{\gamma}}:=\argmin_{\boldsymbol{v} \geq 0} \|\tilde{\boldsymbol{x}}_\gamma - \boldsymbol{A}_{\gamma, S_\gamma} \boldsymbol{v} \|_2^2$.
           \STATE Keep $\tau$ largest entries of $\boldsymbol{w}_\gamma$, zero out the rest.
           \STATE Update $S_{\gamma}$ and $\boldsymbol{w}_{\gamma,S_{\gamma}}$.
           \STATE Update residual: $\boldsymbol{r} \gets \tilde{\boldsymbol{x}}_\gamma - \boldsymbol{A}_{\gamma, S_\gamma} \boldsymbol{w}_{\gamma,S_{\gamma}}$.
       \ENDWHILE
   \ENDFOR
   \STATE Return $\displaystyle\tilde{S} := \bigcup_{\gamma=1}^\Gamma S_\gamma$.
\end{algorithmic}
\end{algorithm}

We integrate the DVP strategy into an MIO solver and refer to the resulting hybrid method as MIO-DVP, parameterized by $\tau_0$. The approach significantly reduces the search space—especially for large $d$ (e.g., $d = 168$ for weekly periodicity)—and makes MIO tractable at scale. MIO-DVP can be viewed as a \textit{backbone-type algorithm} \cite{bertsimas2022backbone}: an iterative two-stage method that first screens candidate features and then solves the final problem over this reduced support set. Such screening strategies have been developed from both statistical (e.g., sure screening \cite{fan2008sure}) and optimization (e.g., safe screening \cite{atamturk2020safe}) perspectives.


\section{Spatially and Temporally-Varying Sparse Autoregression}\label{spatially_varying_model}

In this section, we introduce the Spatially and Temporally-Varying Sparse Autoregression (STV-SAR) model, a generalization of TV-SAR that accounts for both temporal and spatial variations in AR behavior. While TV-SAR captures temporally-varying dynamics within a single univariate time series, STV-SAR is designed for settings involving large spatiotemporal panels of time series (e.g., satellite-based climate data), where thousands of spatial locations exhibit their own local dynamics. This additional spatial dimension leads to a more expressive---but also more complex---model. In particular, when the number of spatial locations (expressed in terms of their latitudes and longitudes) $M \times N = 1$, STV-SAR reduces to TV-SAR. Conversely, STV-SAR enables us to model heterogeneity across locations while leveraging global sparsity and seasonality structures. To solve the associated MIO problem efficiently, we describe a two-stage optimization scheme that separates the learning processes of the global support set and individual coefficient vectors.

\subsection{Model Description}

Considering a collection of time series arranged over an $M \times N$ spatial grid, we let $\boldsymbol{X}_{\gamma,t} \in \mathbb{R}^{M \times N}$ denote the spatial matrix at time $t \in [T_\gamma]$ in time segment $\gamma \in [\Gamma]$. Each grid cell $(m,n)$ contains a multivariate time series $\{x_{m,n,\gamma,t}\}_{t\in[T_\gamma],\gamma\in[\Gamma]}$, in which any $\gamma$th univariate time series is of length $T_{\gamma}$. Following the same logic as in Section \ref{time_varying_model}, we model each time series with a $d$-order AR process, with coefficient vectors $\boldsymbol{w}_{m,n,\gamma} \in \mathbb{R}^d$, and impose a shared global support set across all spatial locations and time segments. Let $\tilde{\boldsymbol{x}}_{m,n,\gamma}$ and $\boldsymbol{A}_{m,n,\gamma}$ be the lagged response vector and design matrix (as in Eq.~\eqref{data_pair}), respectively, then the MIO problem is formulated as follows,
\begin{equation} \label{spatially_time_varying_ar_MIO_problem}
\begin{aligned}
\min_{\Phi,\,\boldsymbol{z}} \quad & \sum_{m,n,\gamma}\|\tilde{\boldsymbol{x}}_{m,n,\gamma}-\boldsymbol{A}_{m,n,\gamma}\boldsymbol{w}_{m,n,\gamma}\|_2^2 \\
\text{s.t.} \quad &
0 \leq \boldsymbol{w}_{m,n,\gamma} \leq \mathcal{M} \cdot \boldsymbol{z}, \quad \forall m,n,\gamma, \\
& \sum_{k=1}^d z_k \leq \tau, \\
& z_k \in \{0,1\}, \quad \forall k \in [d],
\end{aligned}
\end{equation}
where $\Phi=\{\boldsymbol{w}_{m,n,\gamma}\}_{m\in[M],n\in[N],\gamma\in[\Gamma]}$ denotes the set of coefficient vectors. The binary vector $\boldsymbol{z} \in \{0,1\}^d$ encodes the shared support set, $\mathcal{M} > 0$ is a sufficiently large upper bound on the coefficients, and $\tau \in \mathbb{Z}^+$ is a global sparsity budget.

This formulation assumes that only a few autoregressive lags drive the spatiotemporal dynamics across the entire system. For instance, in climate systems, monthly time series such as temperature or precipitation often exhibit strong seasonal structure. By setting $d = 12$, the model can select from lags corresponding to past months in the year. If, for example, $z_{12} = 1$ in the optimal solution, it indicates that the data exhibits strong yearly periodicity—i.e., each month’s value is relevant to its value one year ago. Other selected lags (e.g., $z_1$, $z_3$, etc.) can be interpreted as short-term auto-correlations or sub-seasonal effects.

\subsection{Acceleration via Global Support Estimation} \label{subsec:global-support}

When dealing with large spatiotemporal systems—such as climate datasets covering thousands of grid cells across multiple decades—estimating a separate SAR model for each individual time series becomes computationally infeasible. However, these time series often share common underlying periodicity and auto-correlation patterns. To exploit this structure, we first estimate a global SAR structure by fitting a single sparse coefficient vector $\boldsymbol{w} \in \mathbb{R}^d$ across $M\times N\times \Gamma$ time series. This global sparsity pattern can be reused to simplify subsequent localized coefficient estimation and identify a shared support set that generalizes across space and time. According to the property of matrix trace, we rewrite the objective function as follows,
\begin{equation}
\begin{aligned}
f(\boldsymbol{w}) &= \sum_{m,n,\gamma} \left\| \tilde{\boldsymbol{x}}_{m,n,\gamma} - \boldsymbol{A}_{m,n,\gamma} \boldsymbol{w} \right\|_2^2 \\
&= \sum_{m,n,\gamma} \left( \tilde{\boldsymbol{x}}_{m,n,\gamma} - \boldsymbol{A}_{m,n,\gamma} \boldsymbol{w} \right)^\top \left( \tilde{\boldsymbol{x}}_{m,n,\gamma} - \boldsymbol{A}_{m,n,\gamma} \boldsymbol{w} \right) \\
&= \sum_{m,n,\gamma} \Bigl( \operatorname{tr}\bigl(\boldsymbol{w} \boldsymbol{w}^\top \boldsymbol{A}_{m,n,\gamma}^\top \boldsymbol{A}_{m,n,\gamma}\bigr) \\
&\quad\quad\quad\quad - 2 \boldsymbol{w}^\top \boldsymbol{A}_{m,n,\gamma}^\top \tilde{\boldsymbol{x}}_{m,n,\gamma} \Bigr) + C \\
&= \operatorname{tr}\left( \boldsymbol{w} \boldsymbol{w}^\top \sum_{m,n,\gamma} \boldsymbol{A}_{m,n,\gamma}^\top \boldsymbol{A}_{m,n,\gamma} \right) \\
&\quad\quad\quad\quad - 2 \boldsymbol{w}^\top \sum_{m,n,\gamma} \boldsymbol{A}_{m,n,\gamma}^\top \tilde{\boldsymbol{x}}_{m,n,\gamma} + C,
\end{aligned}
\end{equation}
where $C$ is the constant term. In particular, if one defines the following matrix and vector:
\begin{equation}
\boldsymbol{P} := \sum_{m,n,\gamma} \boldsymbol{A}_{m,n,\gamma}^\top \boldsymbol{A}_{m,n,\gamma}, \quad
\boldsymbol{q} := \sum_{m,n,\gamma} \boldsymbol{A}_{m,n,\gamma}^\top \tilde{\boldsymbol{x}}_{m,n,\gamma}.
\end{equation}
Then the objective can be simplified to
\begin{equation}
f(\boldsymbol{w}) = \operatorname{tr}(\boldsymbol{w} \boldsymbol{w}^\top \boldsymbol{P}) - 2 \boldsymbol{w}^\top \boldsymbol{q} + C.
\end{equation}

Following that form, we encode the sparsity using binary variables $\boldsymbol{z} \in \{0,1\}^d$ and rewrite the MIO as follows,
\begin{equation}
\begin{aligned}
\min_{\boldsymbol{w},\, \boldsymbol{z}} \quad & \operatorname{tr}(\boldsymbol{w} \boldsymbol{w}^\top \boldsymbol{P}) - 2 \boldsymbol{w}^\top \boldsymbol{q} \\
\text{s.t.} \quad & 0 \leq \boldsymbol{w} \leq \mathcal{M} \cdot \boldsymbol{z}, \quad \forall m,n,\gamma, \\
& \sum_{k=1}^d z_k \leq \tau, \\
& z_k \in \{0,1\}, \quad \forall k \in [d].
\end{aligned}
\end{equation}
Thus, the resulting global support set is given by
\begin{equation}\label{support_set}
\Omega := \{k \in [d] \mid w_k > 0 \} = \operatorname{supp}(\boldsymbol{w}) = \operatorname{supp}(\boldsymbol{z}).
\end{equation}

\subsection{Estimating Individual Coefficient Vectors} \label{subsec:indiv-coef}

In practice, learning the global support set is advantageous because (i) the underlying periodicity structure can be quantified across $M\times N$ spatial locations and $\Gamma$ time segments, and (ii) the dominant indices of auto-correlations can be identified and estimated efficiently without having to find them explicitly for each individual coefficient vector. To learn the sparse coefficient vectors $\{\boldsymbol{w}_{m,n,\gamma}\}_{m\in[M],n\in[N],\gamma\in[\Gamma]}$ within the given support set $\Omega$ in Eq.~\eqref{support_set}, the optimization problem now becomes
\begin{equation}
\begin{aligned}
\min_{\boldsymbol{w}_{m,n,\gamma}}\quad &\|\tilde{\boldsymbol{x}}_{m,n,\gamma}-\boldsymbol{A}_{m,n,\gamma}\boldsymbol{w}_{m,n,\gamma}\|_2^2 \\
\text{s.t.}\quad&\mathcal{P}_{\Omega}(\boldsymbol{w}_{m,n,\gamma})\geq0, \quad
\mathcal{P}_{\Omega}^{\perp}(\boldsymbol{w}_{m,n,\gamma})=0, \\
\end{aligned}
\end{equation}
for all $m\in[M],n\in[N],\gamma\in[\Gamma]$. Here, $\mathcal{P}_{\Omega}(\cdot)$ denotes the orthogonal projection supported on $\Omega$. For any vector $\boldsymbol{w}\in\mathbb{R}^{d}$ with entries $\{w_{k}\}_{k\in[d]}$, the orthogonal projection takes $[\mathcal{P}_{\Omega}(\boldsymbol{w})]_{k}=w_k$ if $k\in\Omega$; otherwise, $[\mathcal{P}_{\Omega}^{\perp}(\boldsymbol{w})]_{k}=0$ for any $k\notin\Omega$ in the complement of $\Omega$. That means that the $k$th entry of $\boldsymbol{w}_{m,n,\gamma}$ is zero when satisfying $k\in[d]$ and $k\notin\Omega$ simultaneously. Thus, we only have $|\Omega|$ entries in $\boldsymbol{w}_{m,n,\gamma}$ to estimate. By doing so, we can solve this optimization problem by quadratic optimization with linear constraints. 

\section{Experiments}\label{experiments}

In this section, we evaluate the proposed sparse autoregression (SAR) models through a series of experiments on real-world time series datasets. We consider two distinct domains: urban mobility and climate science. First, we analyze a temporally-varying dataset of ridesharing trips in NYC, which exhibits rich daily and weekly periodicities. This dataset spans multiple years (2019–2023), enabling us to detect how temporal patterns evolve across pre-pandemic, pandemic, and post-pandemic periods. Second, we study large-scale climate datasets, including North American atmospheric variables and global sea surface temperatures. These datasets are both spatially and temporally structured, allowing us to uncover the geographical organization of seasonal effects and trace how climate patterns shift across time.

\subsection{Human Mobility Periodicity}

This case study focuses on identifying and tracking periodic patterns in human mobility using ridesharing data in NYC from 2019 to 2023. Due to daily and weekly commuting cycles and disruptions such as COVID-19, the underlying temporal structure is both periodic and nonstationary. We demonstrate how the proposed TV-SAR model captures this evolving structure and compare alternative solution approaches for model estimation.

\subsubsection{Dataset and Motivation for TV-SAR}

We apply the proposed TV-SAR model to NYC ridesharing data spanning February 2019 to December 2023, focusing on mobility patterns at John F. Kennedy International Airport. Airport-related trips are known to follow strong daily and weekly rhythms due to fixed flight schedules.

\begin{figure*}[h!]
\centering
\subfigure[Pickup trips across weeks]{
    \centering
    \includegraphics[width = 0.4\textwidth]{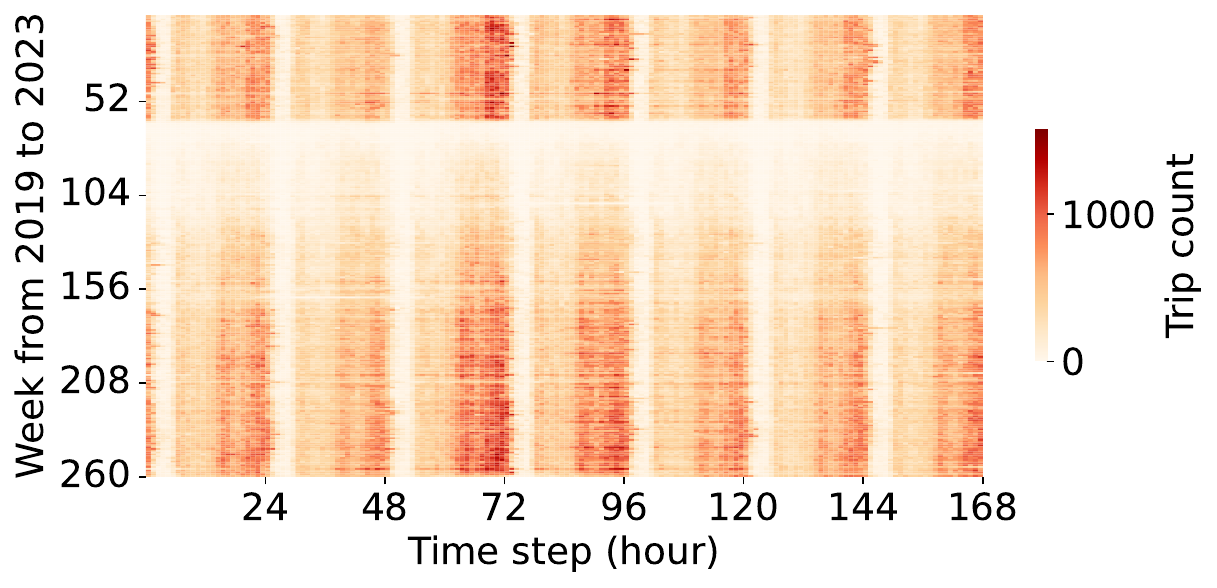}\label{NYC_JFK_airport_ts_pickup}
}
\subfigure[Dropoff trips across weeks]{
    \centering
    \includegraphics[width = 0.4\textwidth]{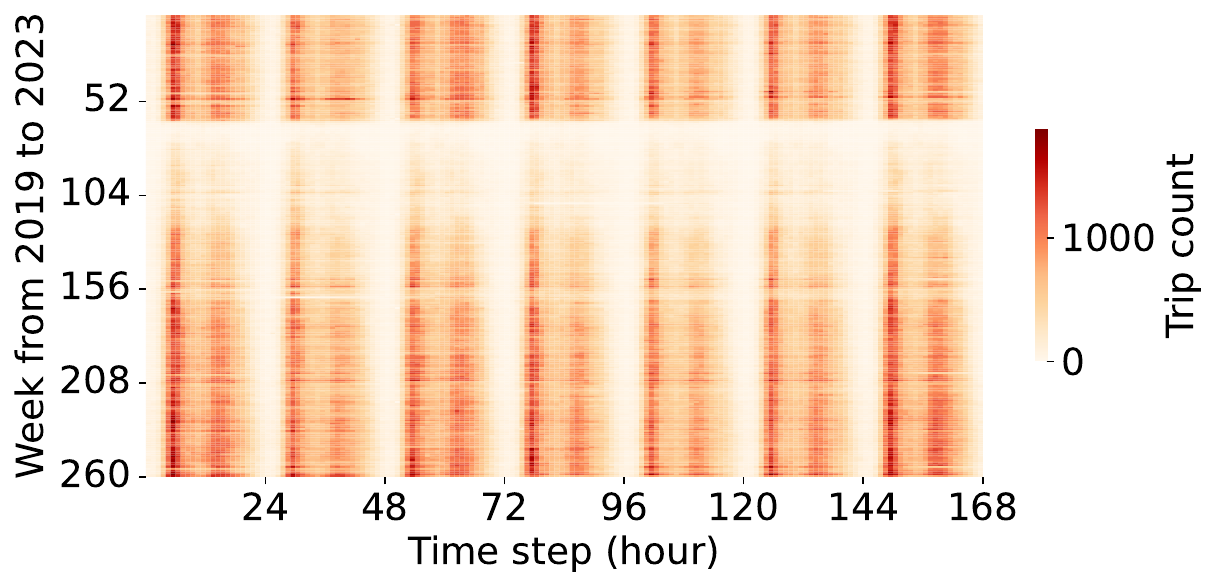}\label{NYC_JFK_airport_ts_dropoff}
}
\subfigure[Aggregated time series of pickup trips]{
    \centering
    \includegraphics[width = 0.4\textwidth]{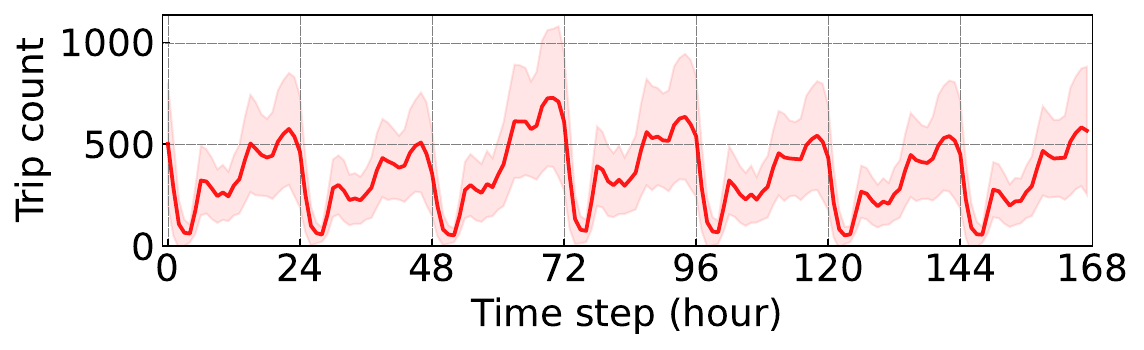}\label{NYC_JFK_airport_ts_pickup_curve}
}
\subfigure[Aggregated time series of dropoff trips]{
    \centering
    \includegraphics[width = 0.4\textwidth]{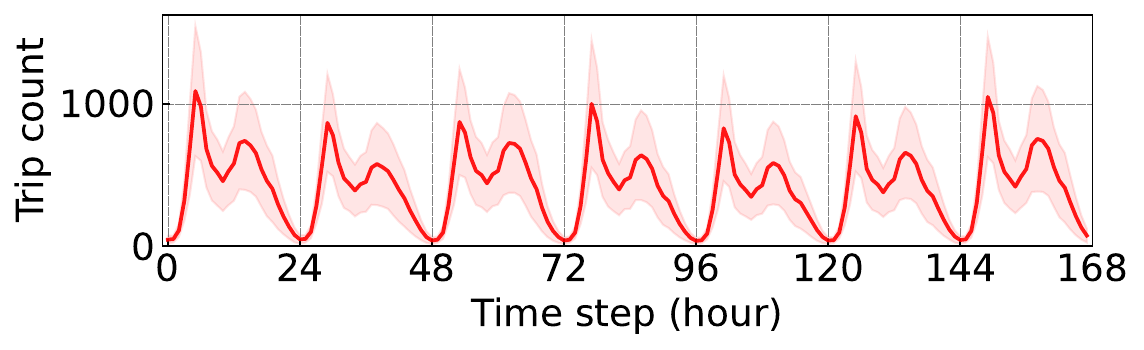}\label{NYC_JFK_airport_ts_dropoff_curve}
}
    \caption{Hourly time series of the aggregated ridesharing trip counts at JFK International Airport from 2019 to 2023. (a-b) The row of each heatmap refers to the ridesharing trip time series of each week. (c-d) The time series refers to the average ridesharing trips of each hour within a week window, while the standard deviations are also presented. 
    }
  \label{NYC_JFK_airport_heatmap}
\end{figure*}

Figures~\ref{NYC_JFK_airport_ts_pickup} and \ref{NYC_JFK_airport_ts_dropoff} visualize the weekly evolution of pickup and dropoff activities, with each heatmap row corresponding to one week (approximately 260 in total). Pickup trips show higher variability due to factors such as flight delays, baggage collection, and access constraints to pickup zones; demand peaks in the evening hours. In contrast, dropoff trips exhibit clearer regularity, with consistent morning and afternoon peaks aligned with flight departures. Aggregated time series in Figures~\ref{NYC_JFK_airport_ts_pickup_curve} and \ref{NYC_JFK_airport_ts_dropoff_curve} further confirm that dropoff trips display stronger daily and weekly periodicity compared to pickups.

Furthermore, as shown in Figure~\ref{NYC_JFK_airport_heatmap}, the ridesharing pickup and dropoff time series display long-term temporal variability. Notably, there is a sharp decline in trip volumes in 2020 due to the COVID-19 pandemic, followed by a gradual recovery through the end of 2021. These observations suggest that the periodic structure of the ridesharing time series is not only rich but also subject to external disruptions. This motivates the use of a temporally-varying model such as TV-SAR to flexibly capture changes in autoregressive patterns across months and years.

\subsubsection{Model Training: Comparison of Estimation Algorithms}

We compare three algorithms for solving Problem~\eqref{optimization_sparse_autoregression} on the ridesharing pickup time series over five years: NNSP, MIO-DVP, and MIO. All methods are tested at two sparsity levels, $\tau = 4$ and $\tau = 6$. For MIO-DVP, we additionally experiment with different pruning thresholds $\tau_0 = 5$ and $\tau_0 = 10$ to examine how the preselected support set size influences solution quality.

Table~\ref{ridesharing_obj_func} reports the objective function values under each setting. As expected, the MIO algorithm consistently achieves the best performance, finding globally optimal solutions. NNSP serves as a baseline and exhibits inferior solution quality, particularly at higher sparsity levels. The performance of MIO-DVP lies in between: with an effective pruning threshold (e.g., $\tau_0 = 10$), it nearly matches the performance of full MIO while incurring significantly lower computational cost.

\begin{table}[htbp]
    \centering
    \caption{Objective function $f(\boldsymbol{w})$ in Eq.~\eqref{optimization_sparse_autoregression} on the ridesharing pickup trip time series at JFK. The solution algorithms include NNSP, MIO-DVP, and MIO. The unit of objective function values is $\times10^7$. Note that the lowest objective function values are emphasized in bold fonts. The last two rows present the average computational times (in seconds). 
    }
    \label{ridesharing_obj_func}
    \small
    \begin{tabular}{c|c|ccccccc}
    \toprule
    \multirow{2}{*}{Data} & \multirow{2}{*}{Sparsity} & \multirow{2}{*}{NNSP} & MIO-DVP & MIO-DVP & \multirow{2}{*}{MIO} \\
    &  &  & ($\tau_0=5$) & ($\tau_0=10$) &  \\
    \midrule
    \multirow{2}{*}{2019} 
    & $\tau=4$ & 8.48 & 8.48 & \textbf{8.24} & \textbf{8.24} \\
    & $\tau=6$ & 8.41 & - & \textbf{8.07} & \textbf{8.07} \\
    \hline
    \multirow{2}{*}{2020} 
    & $\tau=4$ & 2.12 & 2.12 & \textbf{1.90} & \textbf{1.90} \\
    & $\tau=6$ & 2.03 & - & \textbf{1.86} & \textbf{1.86} \\
    \hline
    \multirow{2}{*}{2021} 
    & $\tau=4$ & 3.11 & 3.11 & \textbf{3.06} & \textbf{3.06} \\
    & $\tau=6$ & 3.06 & - & \textbf{2.97} & \textbf{2.97} \\
    \hline
    \multirow{2}{*}{2022} 
    & $\tau=4$ & 6.85 & 6.76 & \textbf{6.49} & \textbf{6.49} \\
    & $\tau=6$ & 6.69 & - & \textbf{6.34} & \textbf{6.34} \\
    \hline
    \multirow{2}{*}{2023} 
    & $\tau=4$ & 8.59 & 8.45 & \textbf{8.14} & \textbf{8.14} \\
    & $\tau=6$ & 8.39 & - & \textbf{7.95} & \textbf{7.95} \\
    \hline    
    \multirow{2}{*}{Cost} 
    & $\tau=4$ & 0.03\,s & 0.32\,s & 0.67\,s & 221\,s \\
    & $\tau=6$ & 0.04\,s & - & 0.66\,s & 223\,s \\
    \bottomrule
    \end{tabular}
\end{table}

This highlights an important tradeoff: although MIO guarantees the best solution, MIO-DVP offers a scalable alternative with much faster runtimes due to its reduced search space. The pruning parameter $\tau_0$ plays a critical role—aggressive pruning can degrade performance, whereas moderate values (like $\tau_0 = 10$) enable a favorable balance between efficiency and accuracy.

\subsubsection{Results: Periodicity of Ridesharing Trips}

To capture the temporally-varying dynamics present in the dataset, we apply the proposed TV-SAR model, treating each month as an independent time segment. The autoregressive order is set to $d = 168$, corresponding to one week of hourly lags. We use a sparsity level of $\tau = 4$ to allow the model to select dominant lags, particularly targeting daily ($k = 24$) and weekly ($k = 168$) periodicities.

\begin{figure*}[h!]
\centering
\subfigure[Pickup trips]{
    \centering
    \includegraphics[width = 0.42\textwidth]{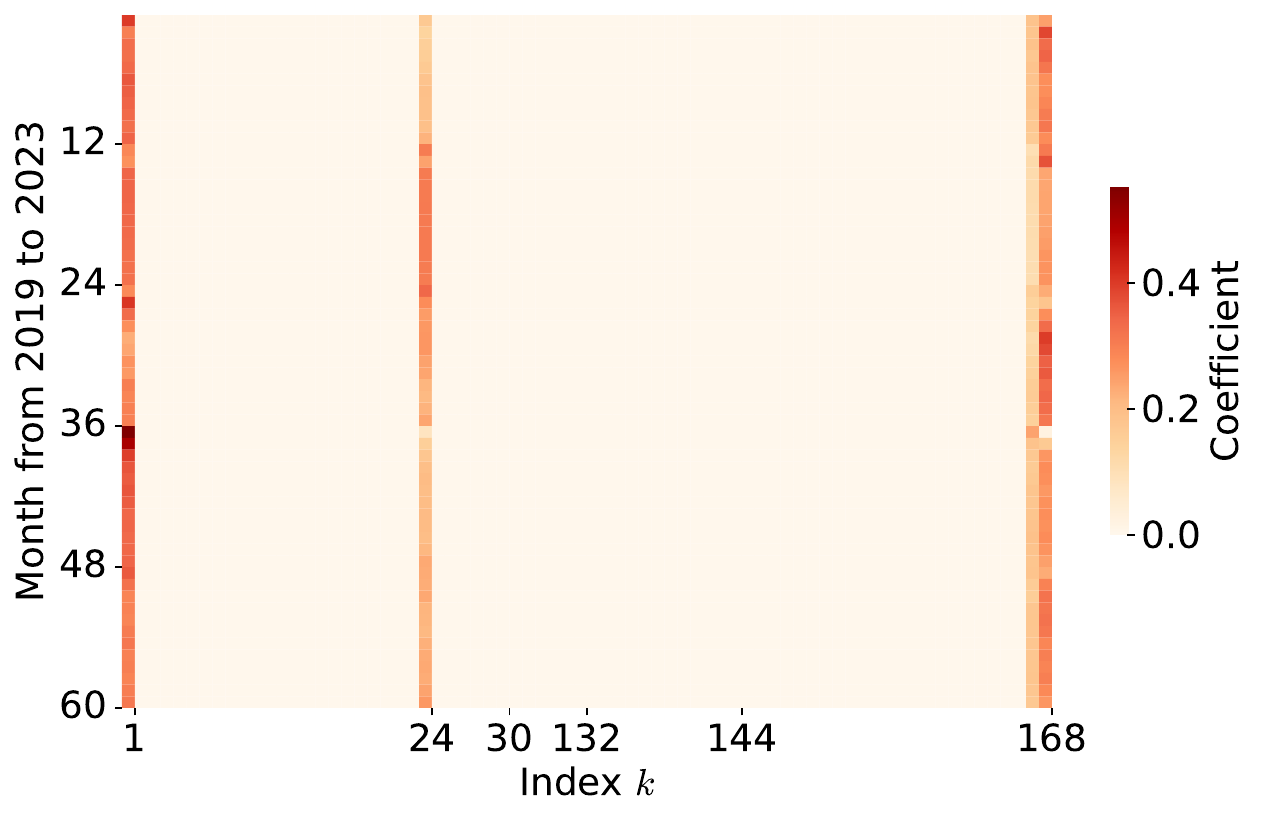}\label{NYC_JFK_pickup_w_tau_3}
}
\subfigure[Dropoff trips]{
    \centering
    \includegraphics[width = 0.42\textwidth]{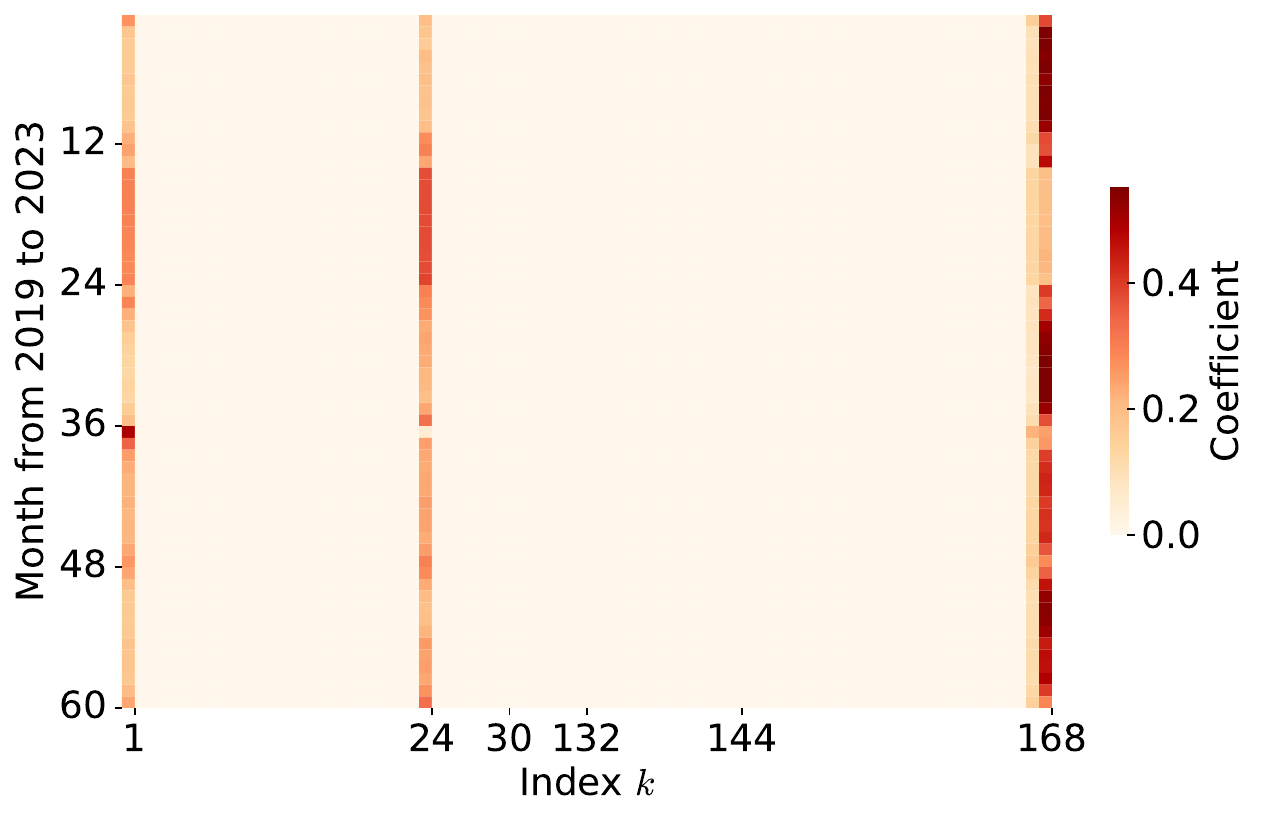}\label{NYC_JFK_dropoff_w_tau_3}
}
\caption{Sparse coefficient vectors $\boldsymbol{w}_{\gamma}\in\mathbb{R}^{168},\,\gamma\in[2,60]$ of TV-SAR on the ridesharing trip time series from February 2019 to December 2023, i.e., 59 months in total. Since each time series data corresponds to to one month, there are 59 coefficient vectors that are represented as 59 rows in the heatmap. Each time series is accumulated from January to the end of the given month. Three most significant auto-correlations are revealed at $k=1,24,168$, referring to local auto-correlation, daily periodicity, and weekly periodicity, respectively.}
  \label{NYC_JFK_w_tau_3}
\end{figure*}

Figure~\ref{NYC_JFK_w_tau_3} shows that the model consistently recovers the support set $\operatorname{supp}(\boldsymbol{w}_{\gamma}) = {1, 24, 167, 168}$ for all $\gamma \in [2, 60]$, reflecting the presence of strong local, daily, and weekly auto-correlations. Figure~\ref{NYC_JFK_pickup_w_tau_3} visualizes how the periodic structure of pickup trips evolves over time. In 2020, the strength of the weekly component decreases remarkably, while the daily component grows stronger. This pattern is similarly observed in the dropoff coefficients (Figure~\ref{NYC_JFK_dropoff_w_tau_3}), where we also note a sharp increase in the local coefficient at $k = 1$ during 2020—suggesting increased short-term variability in dropoff behavior.

Looking across all years, the weekly periodicity at $k = 168$ is generally more pronounced in dropoff trips than in pickup trips. This is consistent with the patterns seen in Figure~\ref{NYC_JFK_airport_heatmap}, where dropoff activity follows more regular flight schedules, while pickups are affected by greater uncertainty (e.g., baggage claim, rideshare wait times). Although periodicity in 2020 differs substantially from the adjacent years, the autoregressive patterns in 2019, 2021, 2022, and 2023 show no significant differences. This indicates that by 2021–2023, human mobility behavior had largely returned to its pre-pandemic structure.

\subsection{Seasonality Patterns of North America Climate Variables}

This case study investigates large-scale seasonal patterns in climate variables across North America using long-term temperature and precipitation data (1980–2019). These data exhibit strong yearly seasonality influenced by geography and climate change, and are structured both spatially and temporally. Our goal is to quantify the strength and variability of these seasonal patterns over space and time. We use the STV-SAR model to uncover the geographical organization of seasonality, track its evolution over decades, and assess the robustness of the learned patterns across spatial resolutions and segmentation schemes.

\subsubsection{Dataset and Motivation for STV-SAR}

\begin{figure*}[ht!]
\centering
\includegraphics[width=1\textwidth]{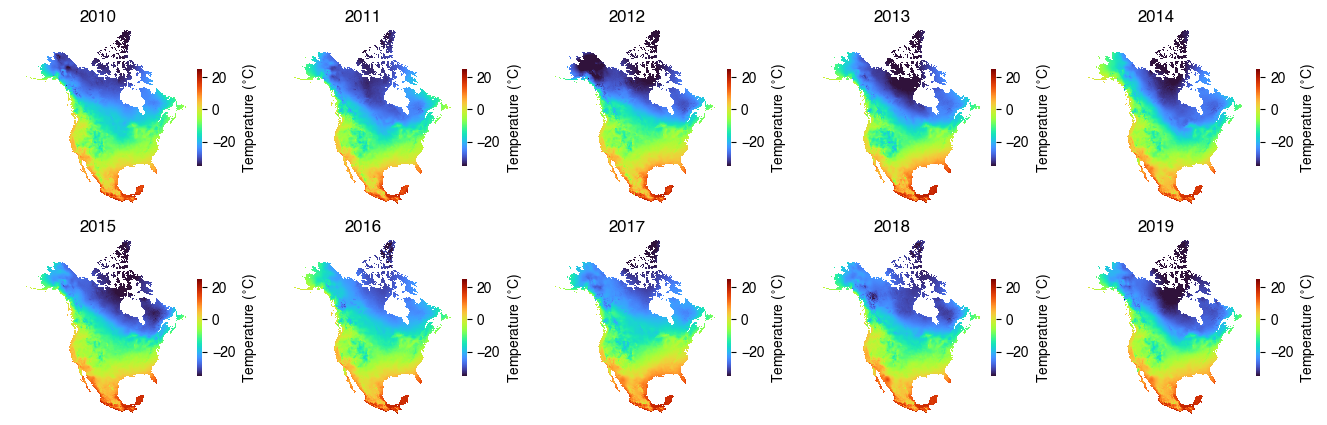}
\caption{Monthly aggregated minimum temperature across North America of January from 2010 to 2019. The color scale represents the temperature in degrees Celsius, with blue tones indicating lower temperatures and red tones indicating higher temperatures.}
\label{daymet_jan_10s}
\end{figure*}

We study the Daymet dataset, which provides monthly climate variables—including minimum/maximum temperature and precipitation—at a high spatial resolution of 1km$\times$1km across North America \cite{thornton2022daymet}.\footnote{\url{https://daac.ornl.gov/DAYMET}} Our analysis focuses on data from 1980 to 2019, resulting in long, spatially-distributed multivariate time series. Figure~\ref{daymet_jan_10s} illustrates the minimum temperature data for January over the past decade. These panels reveal strong seasonal structure and regional climate variability, suggesting the need for models that can capture recurring periodic patterns as well as their changes across space and time.

Given the spatial scale and long-term coverage of this dataset, we aim to (i) quantify yearly seasonality and its regional variation, and (ii) examine how such patterns evolve across decades and spatial resolutions. These goals motivate the use of our proposed STV-SAR model, which is well-suited for discovering global periodic structures in large-scale, multidimensional time series.

\subsubsection{Global Support Verification}


To enable scalable model training on high-resolution spatiotemporal data, we aggregate the original 1\,km$\times$1\,km dataset to a coarser resolution of 10\,km$\times$10\,km. Even at this level, the STV-SAR model of Eq.~\eqref{spatially_time_varying_ar_MIO_problem} must be fit across over 800{,}000 time series. To manage this computational burden, we employ the two-stage optimization scheme outlined in Sections~\ref{subsec:global-support} and \ref{subsec:indiv-coef}. As shown in Table~\ref{computational_cost}, our scheme trains the full model across all time series in under 12 seconds—demonstrating its practical scalability for large-scale climate datasets.

We apply STV-SAR with a prescribed sparsity level $\tau=3$, aiming to capture dominant temporal dependencies that characterize yearly seasonality. For example, using the minimum temperature dataset, we construct monthly time series at each spatial cell over decadal segments (from 1980s to 2010s). Solving the global model yields a common support set $\Omega = \{1,11,12\}$, which includes recent memory ($k=1$), near-annual lags ($k=11,12$), and aligns with expected seasonal dynamics.

\begin{figure}
\centering
\subfigure[2000s]{
\centering
\includegraphics[width=0.46\textwidth]{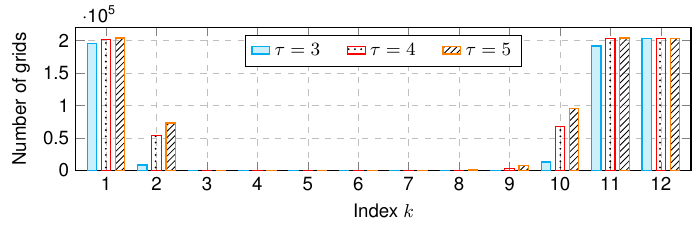}
}
\subfigure[2010s]{
\centering
\includegraphics[width=0.46\textwidth]{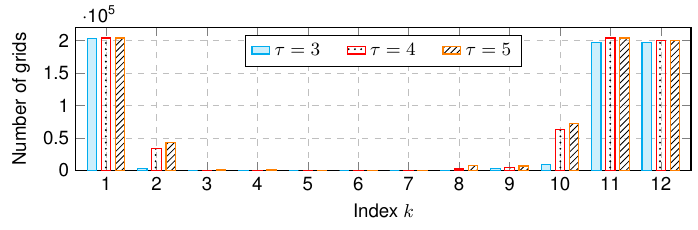}
}
\caption{Support sets of independent SAR models with different sparsity levels tested on the minimum temperature data in 2000s and 2010s across 204,153 grids.}
\label{daymet_supp_set}
\end{figure}

To assess the robustness of this global support set, we also fit independent SAR models to 204,153 time series at 10\,km$\times$10\,km resolution using sparsity levels $\tau \in \{3,4,5\}$. As shown in Figure~\ref{daymet_supp_set}, even without enforcing any global constraint, the most frequently selected lags coincide with the global support set $\Omega$. For $\tau = 3$, the dominant indices are $\{1,11,12\}$; for higher sparsity levels, the support expands to include nearby lags (e.g., $\{2,10\}$), but $\{1,11,12\}$ remain consistently prominent. This confirms that the global support set not only enhances computational efficiency but also meaningfully captures shared temporal structures across space.

\subsubsection{Results: Seasonality Patterns across Four Decades}

\begin{figure*}[ht!]
\centering
\subfigure[Minimum temperature]{
    \centering
    \includegraphics[width=0.9\textwidth]{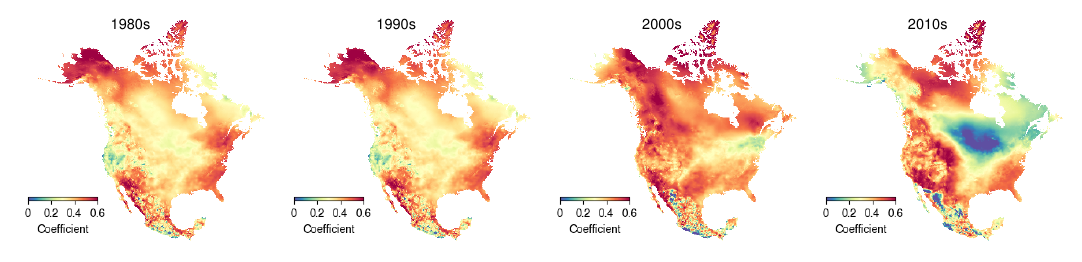}\label{fast_mip_daymet_tmin}
}
\subfigure[Maximum temperature]{
    \centering
    \includegraphics[width=0.9\textwidth]{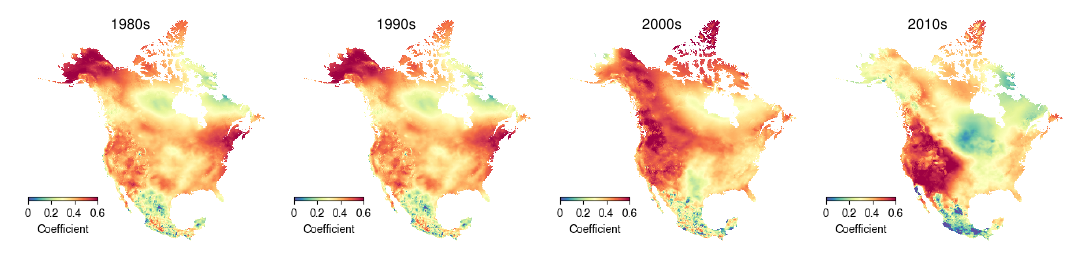}\label{fast_mip_daymet_tmax}
}
\subfigure[Precipitation]{
    \centering
    \includegraphics[width=0.9\textwidth]{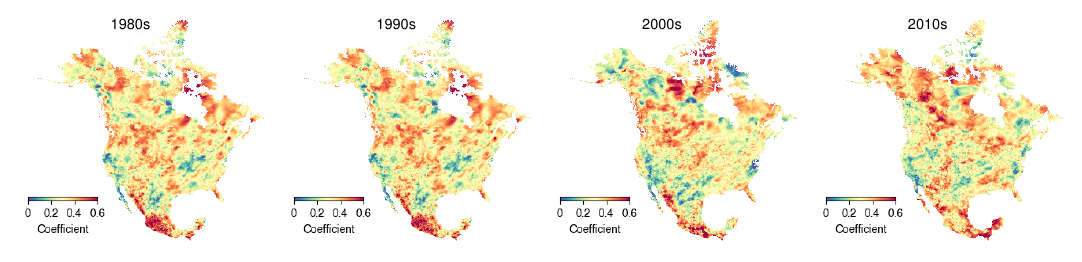}\label{fast_mip_daymet_prcp}
}
\caption{Spatial patterns of the strengths of yearly seasonality quantified by STV-SAR on the North America climate data across the past four decades.}
\label{daymet_w12_sparsity3}
\end{figure*}

Understanding how seasonality patterns evolve over time allows us to quantify the dynamics of large-scale climate systems. Figure~\ref{daymet_w12_sparsity3} visualizes the yearly seasonality of monthly climate time series across North America from 1980 to 2019, using a spatial resolution of 10\,km$\times$10\,km. Higher coefficients correspond to regions with stronger yearly periodicity.

We begin by analyzing the seasonality patterns of minimum and maximum temperature. Across the four decades, areas with stronger yearly seasonality are consistently concentrated in high-latitude regions, such as Canada and Alaska. In contrast, southern regions like Mexico exhibit weaker and more spatially variable seasonality. A notable expansion of highly seasonal regions occurs in the 2000s, suggesting an intensification of temperature-driven seasonal cycles. In the 2010s, two regions stand out: northern Canada (e.g., Nunavut) and the western United States. While the patterns in the 1980s and 1990s are relatively stable, the 2000s and 2010s show substantial shifts—highlighting evolving climate dynamics with potential ecological implications.

Next, we compare seasonal structures across different climate variables. The seasonality of precipitation exhibits significant spatial heterogeneity throughout the decades, even across small neighboring regions. Meanwhile, the seasonality of minimum and maximum temperatures remains relatively stable in the 1980s and 1990s. However, in the 2000s, minimum temperature seasonality intensifies and expands more broadly than that of maximum temperature. In the 2010s, Mexico shows higher seasonality in minimum temperature compared to maximum temperature, whereas the reverse holds in the western U.S. and central North America. Notably, in northern Canada, maximum temperature is less seasonal than minimum temperature. These contrasting patterns suggest differing drivers and sensitivities of seasonal cycles across variables and geographies.

\subsubsection{Sensitivity Analysis: Spatial Resolution}

\begin{figure}[ht!]
\centering
\subfigure[5\,km\,$\times$\,5\,km]{
    \centering
    \includegraphics[width=0.23\textwidth]{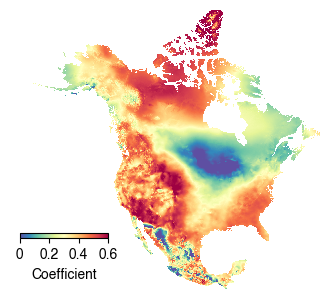}
}
\hspace{-1.2em}
\subfigure[20\,km\,$\times$\,20\,km]{
    \centering
    \includegraphics[width=0.23\textwidth]{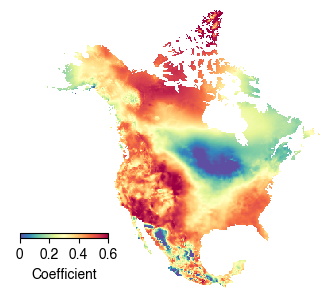}
}
\caption{Spatial patterns of the strengths of yearly seasonality quantified by STV-SAR on the minimum temperature data in 2010s.}
\label{fast_mip_daymet_20km_tmin_w12_sparsity3_10s}
\end{figure}

We now investigate the effect of spatial aggregation by comparing results at three different resolutions: 5\,km$\times$5\,km, 10\,km$\times$10\,km, and 20\,km$\times$20\,km. Figure~\ref{fast_mip_daymet_20km_tmin_w12_sparsity3_10s} illustrates that the spatial distribution of yearly seasonality remains consistent across these resolutions. In particular, the northern regions of Canada, the western United States, and the southeastern United States exhibit strong seasonal signals, while central North America displays substantially weaker seasonality. 

\begin{figure}[ht!]
\centering
\subfigure[5\,km\,$\times$\,5\,km]{
    \centering
    \includegraphics[width=0.23\textwidth]{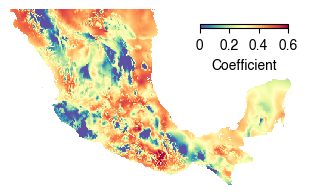}\label{fast_mip_daymet_5km_tmin_mexico_10s}
}
\hspace{-1.2em}
\subfigure[20\,km\,$\times$\,20\,km]{
    \centering
    \includegraphics[width=0.23\textwidth]{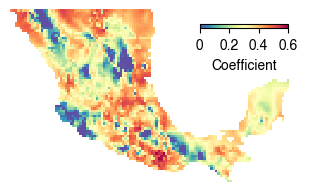}
}
\caption{Spatial patterns of the strengths of yearly seasonality on the minimum temperature data in 2010s within Mexico.}
\label{fast_mip_daymet_mexico_10s}
\end{figure}

Temperature seasonality in Mexico shows high spatial variability and inconsistency among nearby areas, as highlighted in Figure~\ref{fast_mip_daymet_mexico_10s}. Fine-grained resolution (e.g., 5\,km) reveals localized structure that may be smoothed out at coarser scales. This demonstrates that our STV-SAR model captures robust patterns across resolutions while still preserving high-resolution detail where available.

To address the computational challenges of large-scale training, we employ a two-stage optimization scheme that efficiently solves the STV-SAR problem across millions of time series. Table~\ref{computational_cost} reports runtime statistics at each resolution. Remarkably, the model computes over 3 million time series at 5\,km$\times$5\,km resolution within one minute, underscoring the scalability of the proposed method for real-world geospatial applications.

\begin{table}[htbp]
    \centering
    \caption{Runtime of STV-SAR on the North America (minimum) temperature dataset across four decades with different spatial resolutions. Note that $\nu$ denotes the number of time series.}
    \label{computational_cost}
    \small
    \begin{tabular}{l|ccc}
    \toprule
    & 5\,km$\times$5\,km & 10\,km$\times$10\,km & 20\,km$\times$20\,km \\
    \midrule
    Number $\nu$ & 3,343,628 & 816,612 & 196,720 \\
    Runtime (s) & 48.55 & 11.77 & 2.88 \\
    \bottomrule
    \end{tabular}
\end{table}

\subsubsection{Sensitivity Analysis: Phase Segmentation}

\begin{figure*}[ht!]
\centering
\subfigure[From 2000 to 2004]{
    \centering
    \includegraphics[width=0.22\textwidth]{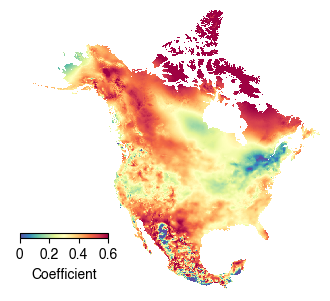}\label{fast_mip_daymet_prcp_5_year_phase_0}
}
\hspace{-1.2em}
\subfigure[From 2005 to 2009]{
    \centering
    \includegraphics[width=0.22\textwidth]{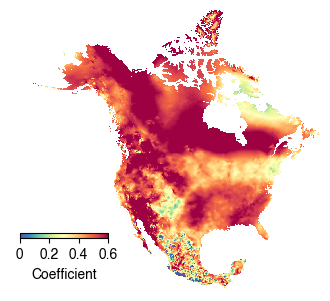}\label{fast_mip_daymet_prcp_5_year_phase_1}
}
\hspace{-1.2em}
\subfigure[From 2010 to 2014]{
    \centering
    \includegraphics[width=0.22\textwidth]{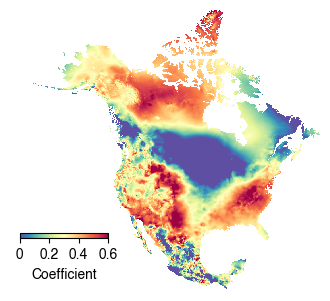}\label{fast_mip_daymet_prcp_5_year_phase_2}
}
\hspace{-1.2em}
\subfigure[From 2015 to 2019]{
    \centering
    \includegraphics[width=0.22\textwidth]{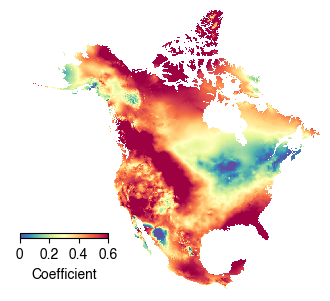}\label{fast_mip_daymet_prcp_5_year_phase_3}
}
\caption{Spatial patterns of the strengths of yearly seasonality quantified by STV-SAR on the minimum temperature data with a 5-year phase (i.e., $T_{\gamma}=60$ months).}
\label{fast_mip_daymet_prcp_5_year}
\end{figure*}

In our main analysis, we segmented the climate time series into decades to capture long-term trends. We now evaluate the sensitivity of our findings to the choice of segmentation granularity by re-estimating seasonality patterns using shorter, 5-year phases on the minimum temperature data from 2000 to 2019. Figure~\ref{fast_mip_daymet_prcp_5_year_phase_0} illustrates the spatial distribution of yearly seasonality from 2000 to 2004, which closely resembles the 1990s pattern shown in Figure~\ref{fast_mip_daymet_tmin}. In the 2005–2009 window (Figure~\ref{fast_mip_daymet_prcp_5_year_phase_1}), seasonality intensifies in several regions, aligning with the dominant pattern of the 2000s. From 2010 to 2014 (Figure~\ref{fast_mip_daymet_prcp_5_year_phase_2}), central North America shows noticeably weaker seasonal signals, whereas the period from 2015 to 2019 (Figure~\ref{fast_mip_daymet_prcp_5_year_phase_3}) marks a rebound in seasonality strength, particularly in Mexico. These subtle temporal shifts are less evident in the decade-level summary (Figure~\ref{daymet_w12_sparsity3}), highlighting the value of finer segmentation for detecting short-term variations in seasonality patterns.

\subsection{Seasonality Patterns of Sea Surface Temperature}

This case study focuses on identifying and tracking yearly periodic patterns in global sea surface temperatures from 1982 to 2019. Sea surface temperature (SST) is a key indicator of climate variability, with dynamics influenced by geography, ocean circulation, and anthropogenic warming. The data exhibit both strong seasonality and long-term trends, making them ideal for evaluating the ability of STV-SAR to capture structured temporal variation at scale. We use the model to uncover geographic patterns of seasonal strength and examine how these patterns evolve across decades and spatial regions.

\subsubsection{Dataset and Motivation for STV-SAR}

We study the NOAA Optimum Interpolation Sea Surface Temperature (OISST) dataset, which provides monthly average temperatures over a $0.25^\circ \times 0.25^\circ$ global grid, comprising $720 \times 1440$ cells (i.e., 1,036,800 spatial locations).\footnote{\url{https://www.ncei.noaa.gov/data/sea-surface-temperature-optimum-interpolation/v2.1/access/avhrr/}} The dataset spans four decades, from January 1982 to December 2019.

\begin{figure}
    \centering
    \includegraphics[width=0.5\textwidth]{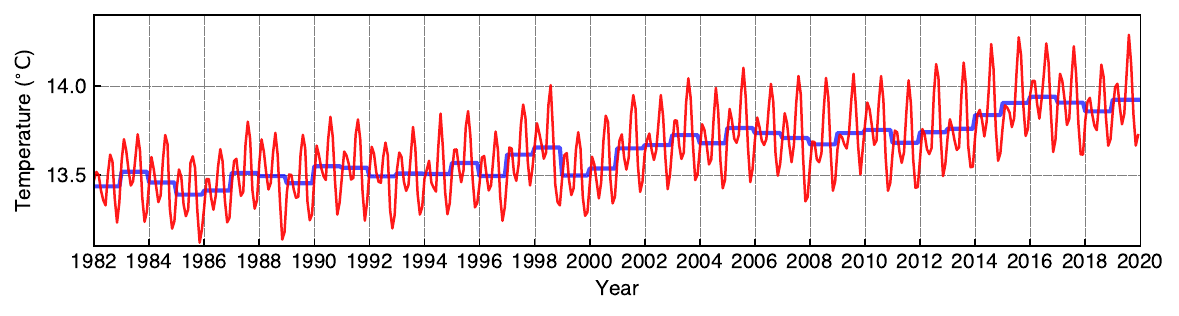}
    \caption{Average values of the monthly sea surface temperature data from January 1982 to December 2019. The average temperature values in 1980s, 1990s, 2000s, and 2010s are 13.46$^\circ\text{C}$, 13.54$^\circ\text{C}$, 13.69$^\circ\text{C}$, and 13.83$^\circ\text{C}$, respectively. The blue and red curves correspond to yearly and monthly average temperature values, respectively.}
    \label{average_temperature_time_series}
\end{figure}

Figure~\ref{average_temperature_time_series} presents both monthly and yearly average sea surface temperatures across this period. The monthly time series reveals strong seasonal fluctuations, while the yearly average trend shows a gradual warming signal consistent with global climate change. These dynamics highlight the importance of quantifying yearly seasonality—not only to understand recurring oceanic patterns but also to characterize the variability and potential instability in the climate system. In particular, stronger seasonality implies more predictable, cyclic dynamics, whereas weaker seasonality can indicate structural shifts or increased variability.

\subsubsection{Results: Spatial and Temporal Structure of Sea Surface Seasonality}

To analyze the temporal structure of sea surface temperatures, we fit the STV-SAR model separately on each decade of data. We fix the model order at $d=12$ and sparsity level $\tau=3$, which allows us to isolate and interpret coefficients corresponding to yearly periodicity.

\begin{figure*}[ht!]
\centering
\subfigure[1980s]{
    \centering
    \includegraphics[width=0.43\textwidth]{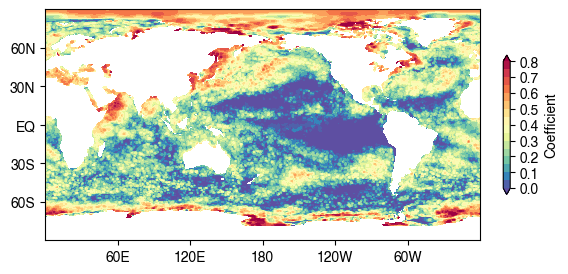}\label{w12_sparsity3_80s}
}
\subfigure[1990s]{
    \centering
    \includegraphics[width=0.43\textwidth]{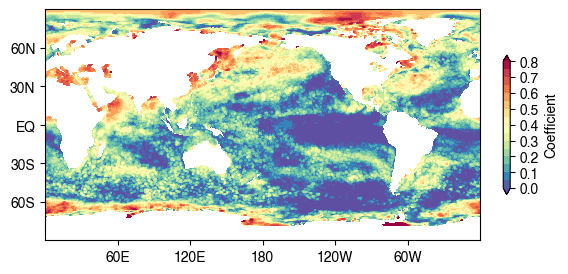}\label{w12_sparsity3_90s}
}
\subfigure[2000s]{
    \centering
    \includegraphics[width=0.43\textwidth]{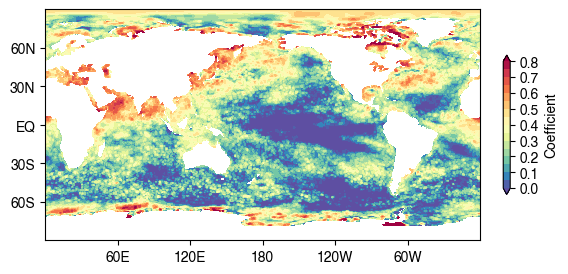}\label{w12_sparsity3_00s}
}
\subfigure[2010s]{
    \centering
    \includegraphics[width=0.43\textwidth]{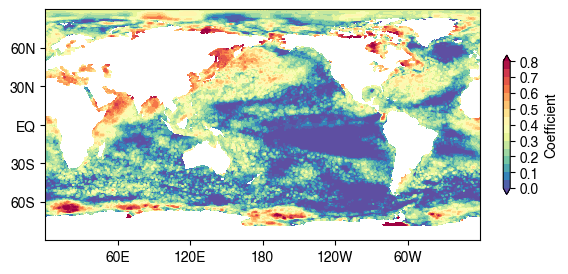}\label{w12_sparsity3_00s}
}
\caption{Spatial patterns of the strengths of yearly seasonality quantified by STV-SAR on the sea surface temperature data in the past four decades.}
\label{w12_sparsity3}
\end{figure*}

Figure~\ref{w12_sparsity3} shows the spatial distribution of the coefficient corresponding to lag 12, which serves as a proxy for yearly seasonality strength. Higher values (in red) indicate regions with stronger temperature seasonality—most notably near Eastern and Central Asia, North America, Europe, and North Africa. In contrast, equatorial regions such as the Pacific Ocean around the El Niño zone exhibit lower coefficients, reflecting weaker seasonality and more irregular variation.

\begin{figure*}[ht!]
\centering
\subfigure[1980s]{
    \centering
    \includegraphics[width=0.33\textwidth]{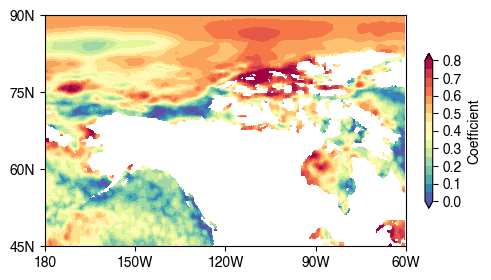}\label{w12_sparsity3_80s_canada}
}\hspace{-0.9em}
\subfigure[1990s]{
    \centering
    \includegraphics[width=0.33\textwidth]{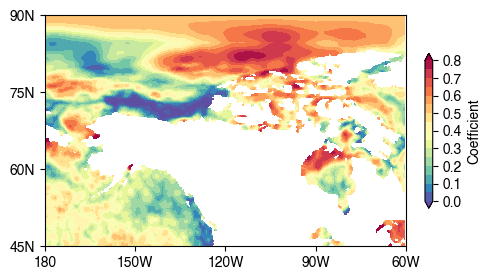}\label{w12_sparsity3_90s_canada}
}\hspace{-0.9em}
\subfigure[2010s]{
    \centering
    \includegraphics[width=0.33\textwidth]{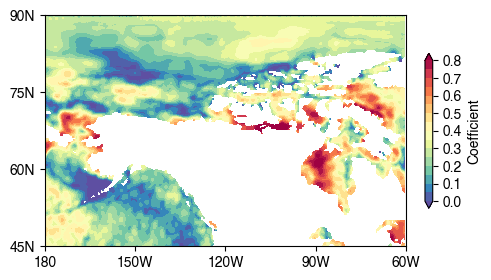}\label{w12_sparsity3_10s_canada}
}
\caption{Spatial patterns of the strengths of yearly seasonality on the sea surface temperature data with highlighted areas around Canada.}
\label{w12_sparsity3_canada}
\end{figure*}

Despite the broad stability of global seasonal patterns across decades, we observe subtle temporal changes. For example, as shown in Figure~\ref{w12_sparsity3_canada}, the regions near Canada and the Arctic Ocean exhibit decreasing seasonality over time. This decline may signal evolving oceanic dynamics due to changes in ocean circulation, warming trends, or atmospheric variability. The Arctic Ocean, in particular, appears to have become notably less seasonal over the past two decades, potentially reflecting broader climate-induced disruptions in polar ocean systems.

\section{Conclusion}\label{conclusion}

This work introduces a unified SAR framework for quantifying periodicity in real-world spatiotemporal time series. We develop interpretable extensions—TV-SAR and STV-SAR—that incorporate temporally and spatially-varying sparsity constraints to reveal structured periodic components. To address the computational challenges posed by the underlying MIO problems, we propose scalable algorithmic solutions: a DVP-based method for TV-SAR and a two-stage optimization scheme for STV-SAR. Applied to both ridesharing mobility data and climate time series, our models successfully uncover meaningful, interpretable seasonal patterns and periodic structures. The learned spatial and temporal dynamics illustrate the practical relevance of periodicity quantification for analyzing complex systems.

We point out the following directions for future improvement. The global support estimation strategy used in STV-SAR is particularly effective when temporal structures are shared across time series, as in climate datasets (e.g., Figure~\ref{daymet_supp_set}). However, for heterogeneous datasets with diverse or complex dynamics, the approximation introduced by the two-stage optimization may lead to estimation bias. Addressing this gap—potentially by integrating adaptive or hierarchical modeling techniques—could further enhance the robustness and generality of the framework.

\ifCLASSOPTIONcompsoc
  \section*{Acknowledgments}
\else
  \section*{Acknowledgment}
\fi

This research is based upon work supported by the U.S. Department of Energy’s Office of Energy Efficiency and Renewable Energy (EERE) under the Vehicle Technology Program Award Number DE-EE0009211 and and DE-EE0011186. The Mens, Manus, and Machina (M3S) is an interdisciplinary research group (IRG) of the Singapore MIT Alliance for Research and Technology (SMART) center.

\ifCLASSOPTIONcaptionsoff
  \newpage
\fi



%

\bibliographystyle{IEEEtran}
\bibliography{references}

%


\begin{IEEEbiography}[{\includegraphics[width=1in,height=1.25in,clip,keepaspectratio]{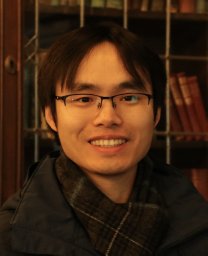}}]{Xinyu Chen} is a Postdoctoral Associate at Massachusetts Institute of Technology, Cambridge, MA, United States. He received his Ph.D. degree from the University of Montreal, Montreal, QC, Canada. His current research centers on machine learning, spatiotemporal data modeling, intelligent transportation systems, and urban science. \end{IEEEbiography}

\begin{IEEEbiography}[{\includegraphics[width=1in,height=1.25in,clip,keepaspectratio]{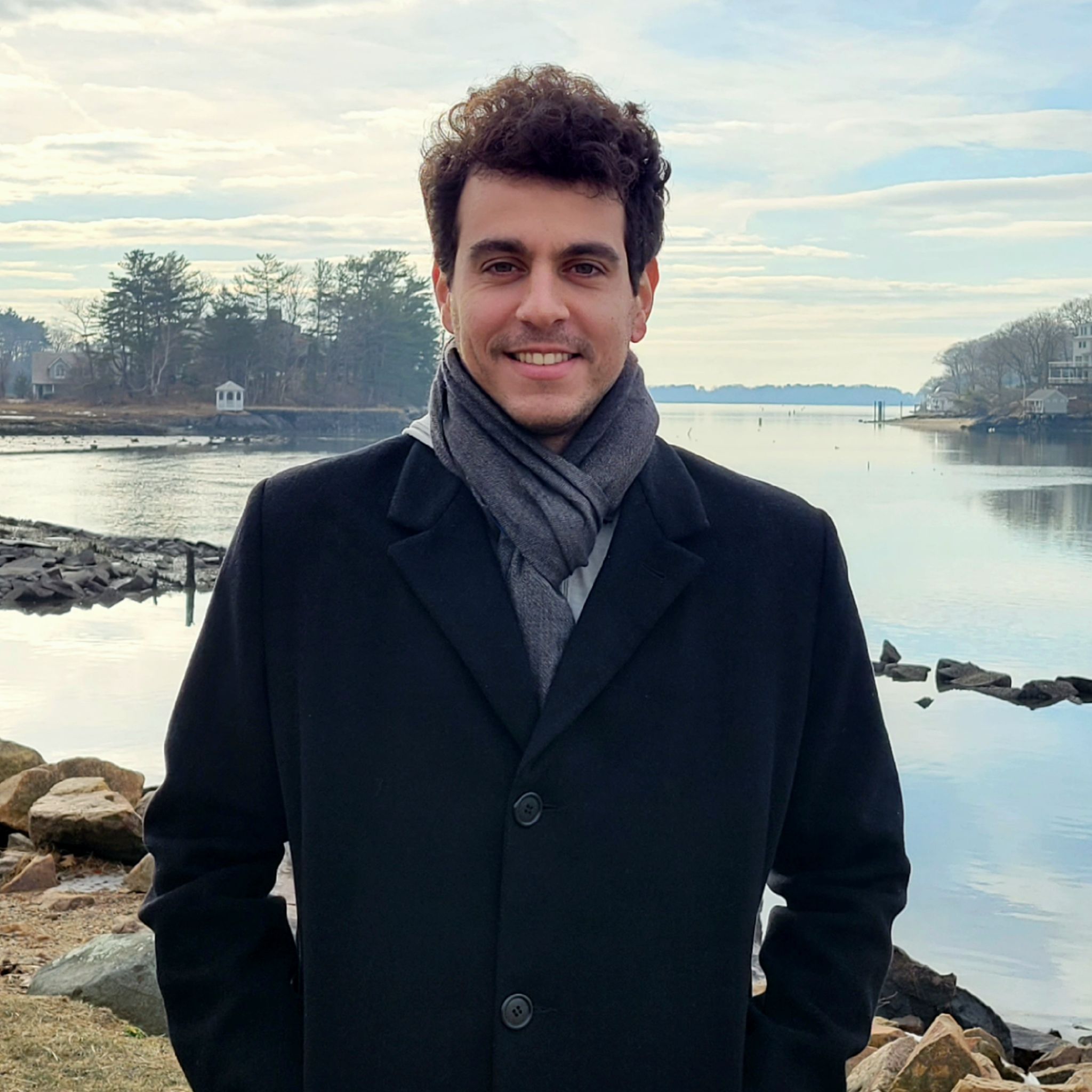}}]{Vassilis Digalakis Jr} is an Assistant Professor of Operations \& Technology Management at the O\&T Management Department at Boston University's Questrom School of Business. He received his Ph.D. in Operations Research at the Operations Research Center at MIT. His research develops methods in trustworthy AI and machine learning, with applications in healthcare and sustainability.
\end{IEEEbiography}

\begin{IEEEbiography}[{\includegraphics[width=1in,height=1.25in,clip,keepaspectratio]{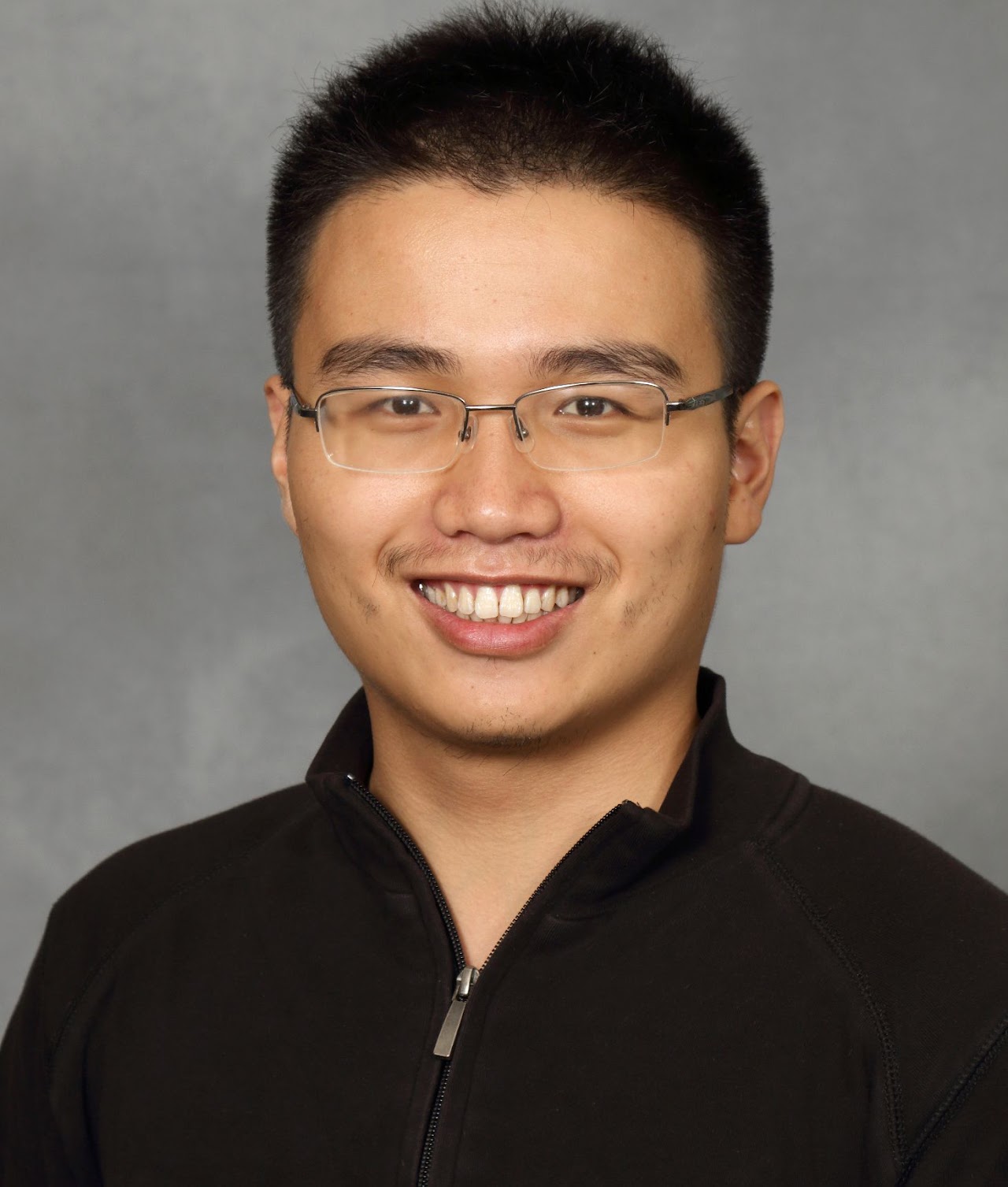}}]{Lijun Ding} is an Assistant Professor at the Department of Mathematics at the University of California, San Diego. He earned his PhD in Operations Research from Cornell University in 2021. He specializes in continuous optimization, especially in semidefinite programming and applications of optimization in statistics.
\end{IEEEbiography}

\begin{IEEEbiography}[{\includegraphics[width=1in,height=1.25in,clip,keepaspectratio]{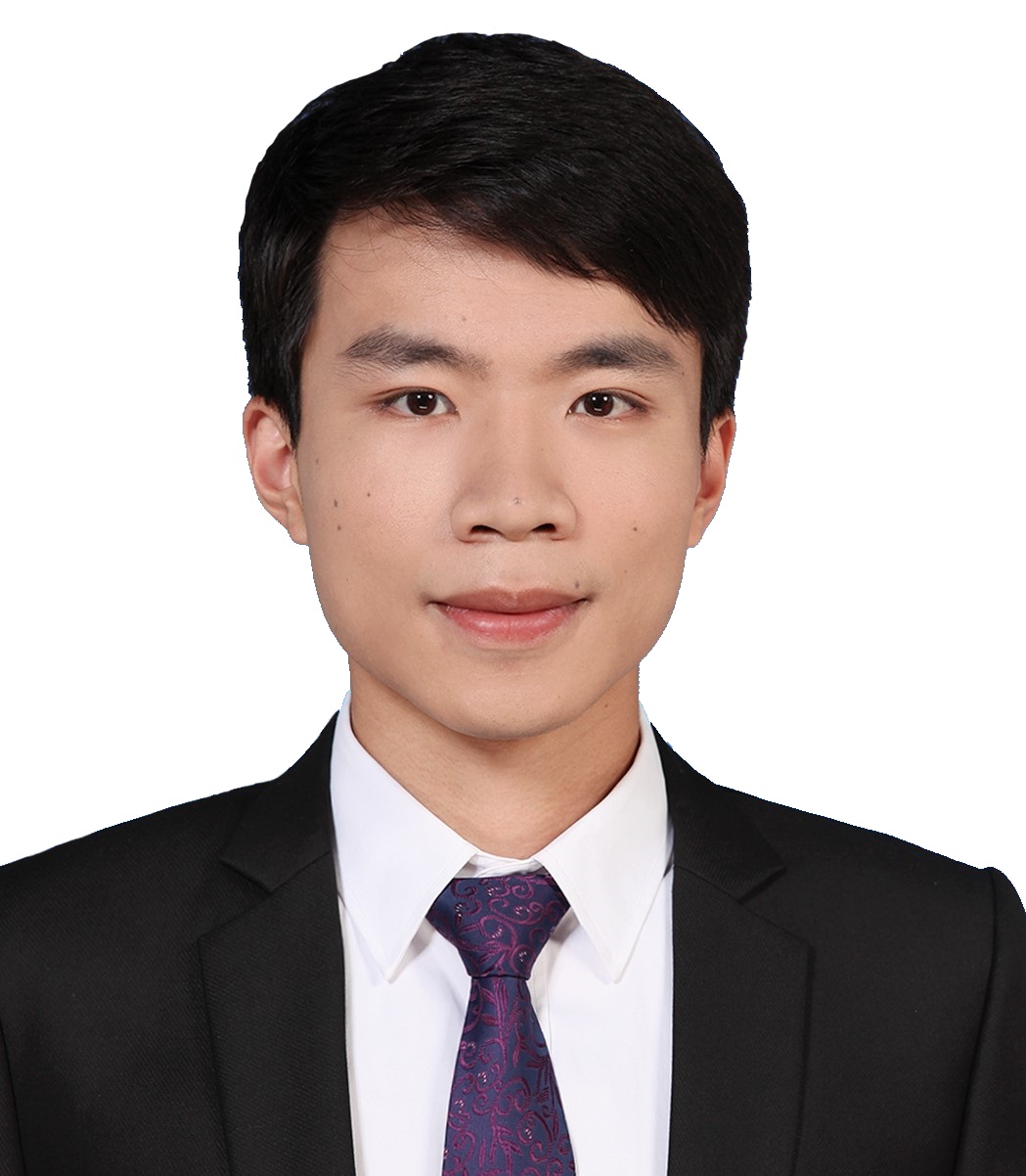}}]{Dingyi Zhuang} is a Ph.D. student in Transportation Engineering at MIT Urban Mobility Lab. He received a B.S. degree in Mechanical Engineering from Shanghai Jiao Tong University in 2019, and an M.Eng. degree in Transportation Engineering from McGill University, in 2021. His research interests lie in deep learning, urban computing, and spatiotemporal data modeling.
\end{IEEEbiography}

\begin{IEEEbiography}[{\includegraphics[width=1in,height=1.25in,clip,keepaspectratio]{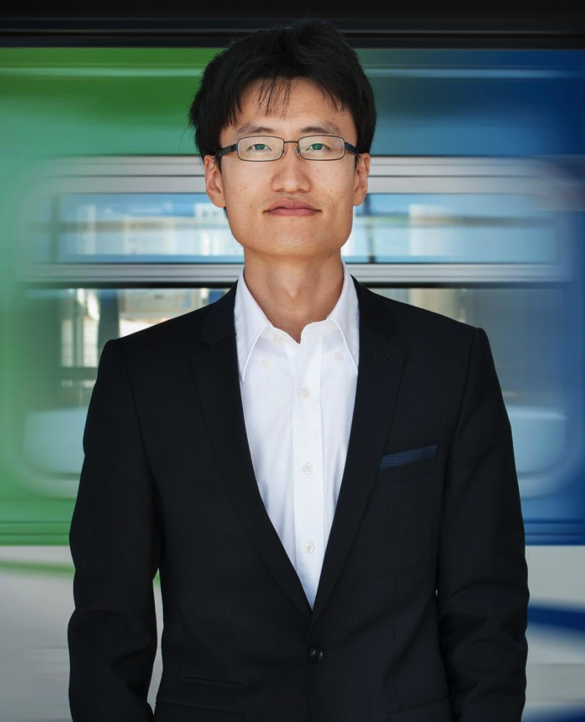}}]{Jinhua Zhao}
is currently the Professor of Cities and Transportation at MIT. He brings behavioral science and transportation technology together to shape travel behavior, design mobility systems, and reform urban policies. He directs the MIT Urban Mobility Laboratory and Public Transit Laboratory.
\end{IEEEbiography}


\vfill


\end{document}